\pdfoutput=1

\documentclass[11pt]{article}

\usepackage[final]{EMNLP/acl}

\newcommand{\cmark}{\text{\ding{51}}}
\newcommand{\xmark}{\text{\ding{55}}}
\newcommand{\qmark}{\text{\sffamily ?}}
\newcommand{\sys}[0]{\textsc{QSpec}}
\usepackage[utf8]{inputenc} 
\usepackage[T1]{fontenc}    
\usepackage{hyperref}       
\usepackage{url}            
\usepackage{booktabs}       
\usepackage{amsfonts}       
\usepackage{nicefrac}       
\usepackage{microtype}      
\usepackage{xcolor}         
\usepackage{graphicx}
\usepackage{multirow}       
\usepackage{array}          
\usepackage{color}
\usepackage{pifont}         
\usepackage{siunitx}        %
\usepackage{amsmath}        %
\usepackage{caption}
\usepackage{mathtools}
\usepackage{enumitem}
\usepackage[table]{xcolor}
\usepackage{makecell} 
\usepackage{float}

\usepackage{times}
\usepackage{latexsym}

\usepackage[T1]{fontenc}

\usepackage[utf8]{inputenc}

\usepackage{microtype}

\usepackage{inconsolata}

\usepackage{graphicx}

\newcolumntype{t}[1]{>{\centering\arraybackslash}m{#1}}
\newcolumntype{C}[1]{>{\centering\arraybackslash}m{#1}}

\title{\sys{}: Speculative Decoding with Complementary Quantization Schemes}

\author{
  Juntao Zhao,
  Wenhao Lu,
  Sheng Wang,
  Lingpeng Kong,
  and Chuan Wu
  \\
  \\
  \texttt{\{juntaozh, whlu, u3009618\}@connect.hku.hk},
  \texttt{\{lpk, cwu\}@cs.hku.hk}
}

\begin{document}
\maketitle
\begin{abstract}

Quantization is widely adopted to accelerate inference and reduce memory consumption in large language models (LLMs). While activation-weight joint quantization enables efficient low-precision decoding, it suffers from substantial performance degradation on multi-step reasoning tasks. We propose \sys{}, a novel quantization paradigm that decouples efficiency from quality by integrating two complementary schemes via speculative decoding: low-precision joint quantization for fast drafting and high-precision weight-only quantization for accurate verification. \sys{} reuses both weights and KV cache across stages, enabling near-zero-cost switching without retraining or auxiliary models. Compared to high-precision baselines, \sys{} achieves up to $1.64\times$ speedup without quality degradation, and outperforms state-of-the-art speculative decoding methods by up to $1.55\times$ in batched settings. Furthermore, \sys{} supports plug-and-play deployment and generalizes well across model scales, quantization methods, and workloads. These properties make \sys{} a practical and scalable solution for high-fidelity quantized LLM serving under memory-constrained scenarios.
Our code is available at \url{https://github.com/hku-netexplo-lab/QSpec}.

\end{abstract}

\section{Introduction} \label{sec: intro} 
Large language models (LLMs) have demonstrated remarkable abilities across various domains, including mathematics, coding, and planning~\citep{ds_math, ds_code, UnderstandingThePlanning_Huang2024}. 
Nonetheless, their immense scales pose substantial challenges for deployment due to high memory and computational demands, especially in resource-limited scenarios (\textit{e.g.}, inference on edge devices). Quantization has been an effective compression technique to facilitate 
LLM inference with limited resources~\citep{awq, quarot, Atom, qserve}. 
By converting high-precision values (\textit{e.g.}, FP16) into their lower-precision counterparts (\textit{e.g.}, INT4), quantization effectively lowers memory and computational requirements, allowing for larger serving batches and model sizes. Furthermore, the reduced memory footprint boosts token generation throughput by accelerating the typically memory-bound autoregressive decoding process~\citep{zhao2024llmpqservingllmheterogeneous}.

\begin{figure}[t]
  \centering
  \includegraphics[width=1\columnwidth]{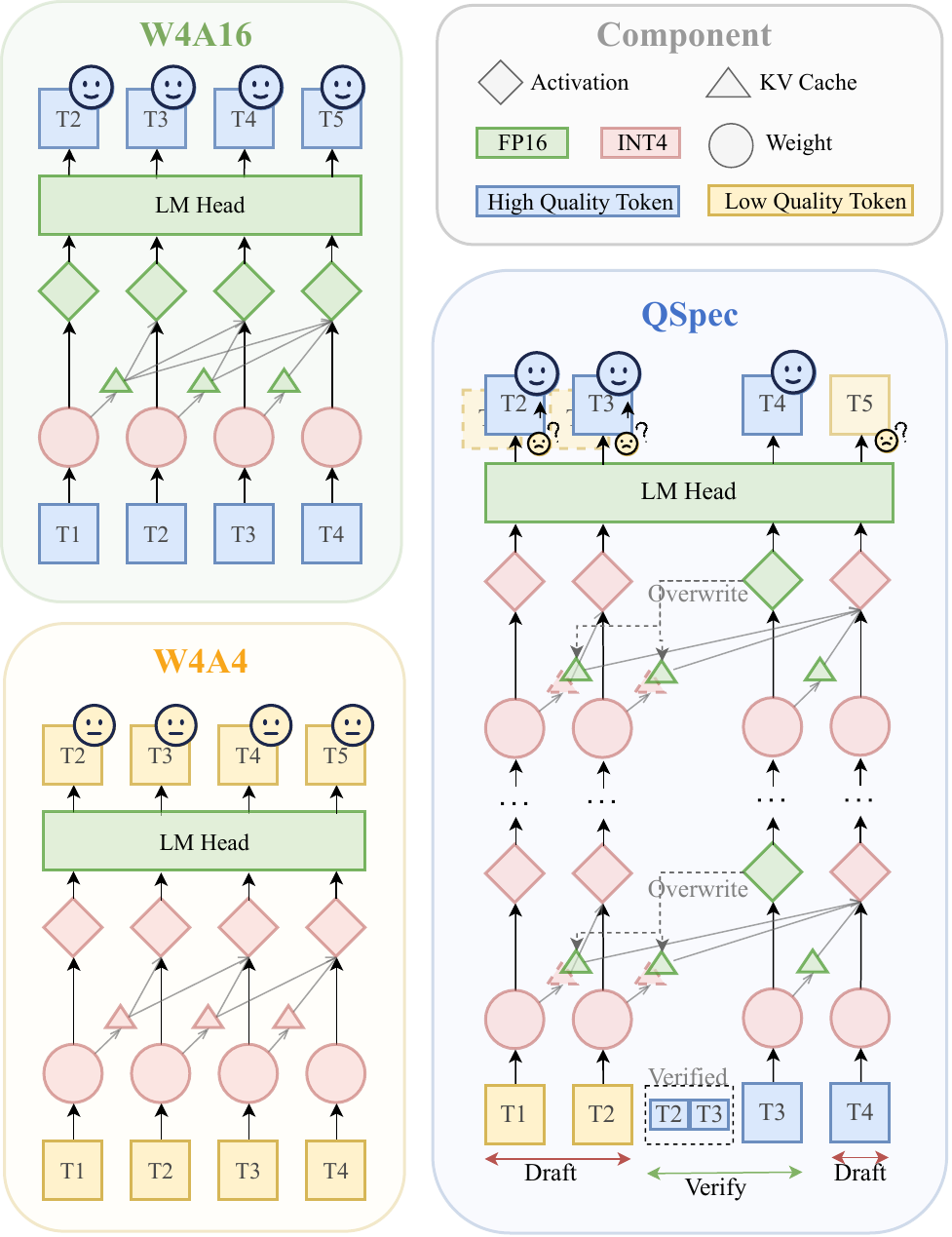}
  \vspace{-4mm}
  \caption{Diagrams of different 4-bit quantization schemes. 
    \textbf{W4A16:} uses 4-bit weight and 16-bit activation for inference. \textbf{W4A4:} further adopts 4-bit activation to utilize low-precision W4A4 kernels. \textbf{\sys{}:} accelerates W4A16 by drafting tokens with W4A4 and verifying them with W4A16, and applies KV cache overwriting for consistent memory consumption. 
  }
  
  \label{fig:qspec_design}
  \vspace{-4mm}
\end{figure}

Based on the quantized objects, recent quantization algorithms can be broadly classified into two categories: weight-only and WXAX: 
(1) Weight-only quantization, represented by W4A16~\citep{awq}, quantizes model weights to low precision (\textit{e.g.}, 4-bit) for storage, 
and then dequantizes them to a higher precision (\textit{i.e.}, FP16) during inference;
(2) WXAX methods, such as W4A4~\citep{quarot, Atom} and W8A8~\citep{xiao2023smoothquant}, simultaneously quantize both weights and activations, and leverage low-precision hardware support for faster execution without dequantizing them to higher precision. 
Nevertheless, WXAX schemes generally suffer model performance degradation due to more low-precision activations used (as verified in Sec.~\ref{sec: motivation}). This poses a tough trade-off between efficacy and efficiency, raising the question:

\emph{``Is there a quantization solution that boosts efficiency while avoiding efficacy degradation?''}.



Considering the comparable performance claims on recent W4A4 methods~\citep{Atom, quarot}, we first contend that their conclusions are biased due to limited evaluation tasks, and W4A4 still experiences significant performance drops when compared to their higher-precision activation counterparts.
Specifically, while W4A4 schemes such as Atom~\citep{Atom} and QuaRot~\citep{quarot} perform well on general tasks, such as PIQA~\citep{piqa}, Winogrande~\citep{ai2:winogrande} and ARC~\cite{allenai:arc}, they demonstrate notable performance declines in multi-step reasoning, 
particularly on mathematical and coding benchmarks~\citep{xiong2024buildingmathagentsmultiturn, guo2024can} (shown in Table~\ref{tab:metric_q}).
This raises concerns about the comprehensiveness of evaluation and emphasizes the necessity of incorporating multi-step reasoning tasks into quantization assessment. 

To answer the above question, we draw inspiration from speculative decoding ~\citep{leviathan2023fastinferencetransformersspeculative, googlespec}, which combines rapid drafting of a small model with high-fidelity verification from a larger model to boost throughput without sacrificing output quality. We propose a novel paradigm called \sys{}, which employs complementary mixed-precision quantization execution to 
deliver \textbf{flawless acceleration} for high precision quantization. (\textit{i.e.}, \textbf{efficiency improvements} that preserve 
the fidelity and memory overhead of high-precision quantization).


Our key insight is that a single weight-quantized model can losslessly toggle between two parallel activation modes: a fast, low-precision mode for drafting (\textit{e.g.}, W4A4) and a high-precision mode for verification (\textit{e.g.}, W4A16). As validated in Sec.~\ref{sec:salient_token}, the token generation with different modes is highly similar. This allows us to adopt a lightweight `draft-verify' scheme, as shown in Figure~\ref{fig:qspec_design}, where tokens drafted with W4A4 are selectively accepted by W4A16 with negligible switching costs. Unlike traditional speculative decoding that requires a draft model, \sys{} shares weights and KV cache across activation modes, achieving zero extra memory overhead.

We evaluate \sys{} on a range of model sizes, quantization methods, and batch sizes. Compared to W4A16, \sys{} offers up to $1.64\times$ higher token generation throughput while preserving fidelity. \textbf{It also effectively compensates for up to 51.11\% quality loss observed in W4A4 on challenging multi-step reasoning tasks such as MATH~\citep{MATH}.} Furthermore, \sys{} requires no training and can be directly integrated into existing inference pipelines.







Our main contributions are summarized as follows:
\vspace{-3mm}
\begin{enumerate}[label=\textbullet, itemsep=0pt, leftmargin=*]

\item We demonstrate that multi-step reasoning tasks are more sensitive to quantization-induced quality degradation than standard benchmarks, and advocate their inclusion for more comprehensive evaluation.

\item We validate and instantiate the feasibility of switching between two quantization schemes of a shared weight-quantized model, as well as their high token-level similarities, illuminating future development of quantization schemes.

\item We introduce \sys{}, the first quantization paradigm that decouples efficiency from quality by combining complementary quantization schemes through speculative decoding with shared weights and KV cache.

\item Experiments across various models and tasks reveal that \sys{} achieves up to $1.64\times$ acceleration without quality loss, making it well-suited for high-fidelity deployment in memory-constrained scenarios.

\end{enumerate}

\section{Motivation}\label{sec: motivation}

\subsection{Compromised Performance of Activation Quantization}\label{sec:reason_better}



State-of-the-art (SOTA) activation-weight joint quantization methods, like Atom~\citep{Atom} and QuaRot~\citep{quarot}, 
achieve notable speed-ups with negligible performance loss compared to weight-only ones. However, we argue that this conclusion is skewed by limited evaluation benchmarks, which fail to capture the negative impacts of activation quantization.

\begin{table}[t]
    \centering
    \caption{Performance of Atom-based quantization schemes with different weight and activation precision across diverse tasks. ``Acc'', ``PPL'' and ``EM'' stand for accuracy, perplexity, and exact match, respectively, with arrows indicating their positive trends. ``W16A16'' refers to standard FP16 inference, where both weights and activations are represented in FP16 precision.} 
    \vspace{-6pt}
    \label{tab:metric_q}
    \resizebox{\linewidth}{!}{
    \begin{tabular}{ccccc}
    \toprule
    \multirow{2}{*}{Task} & \multirow{2}{*}{Metric}    & \multirow{2}{*}{W16A16}   & \multicolumn{2}{c}{Quantization}  \\
    \cmidrule(lr){4-5}
                                   &                                        &                & Atom (W4A16)                 & Atom (W4A4)  \\
    \midrule
    WikiText-2                     & PPL $\downarrow$                       & 7.73                  & 7.87 \textcolor{black}{(+0.15\%)}         & 8.58 \textcolor{black}{(+0.85\%)}  \\
    PIQA (10-shot)                 & EM $\uparrow$                         & 78.6                 & 77.5 (-1.40\%)                 & 75.6 (-3.81\%) \\
    MBPP (0-shot)                  & EM $\uparrow$                         & 42.0                 & 41.5 \textcolor{black}{(-1.19\%)}        & 30.5  \textcolor{red}{(-27.38\%)}  \\
    GSM8K (8-shot)                 & EM  $\uparrow$                         & 79.0                 & 73.4 \textcolor{red}{(-7.09\%)}        & 54.2  \textcolor{red}{(-31.39\%)}  \\
    \bottomrule
    \end{tabular}%
    }
    \vspace{-12pt}
\end{table}




To substantiate this claim, we conduct experiments on Llama-3-8B-Instruct models~\citep{llama3} quantized with W16A16, W4A16, and W4A4 across four benchmarks: PIQA~\citep{piqa}, WikiText-2~\citep{wikitext2}, GSM8K~\citep{gsm8k}, and MBPP~\citep{MBPP}. While PIQA and WikiText-2 are commonly used in quantization evaluation, GSM8K and MBPP involve multi-step reasoning, which remains underexplored in the quantization context despite its great importance. Detailed descriptions of the benchmarks are provided in Appendix~\ref{app:datasets}.

As listed in Table~\ref{tab:metric_q}, 
Atom-based quantization schemes show comparable performance to W16A16 across commonly adopted tasks such as on PIQA and WikiText-2, aligning with the claims in \citet{Atom}. 
However, W4A4 suffers a nearly 30\% average performance decline on complex reasoning tasks (\textit{i.e.}, MBPP and GSM8K), whereas W4A16 only experiences about 4\%.
This indicates that activation quantization leads to several times more performance degradation on multi-step reasoning tasks, despite the improved efficiency.
Besides, the performance trend observed on multi-step reasoning tasks shows a stronger correlation with quantization precision than perplexity does, validating their adequacy in evaluation. 


In summary, activation quantization still incurs significant performance loss on more advanced multi-step reasoning tasks. This necessitates the inclusion of them in quantization evaluation 
for a more comprehensive assessment. Furthermore, this also underscores the demand for a quality-preserving yet efficient quantization paradigm.



\subsection{High-Similarity Token Predictions} \label{sec:salient_token}

\begin{figure}[H]
    \centering
    \includegraphics[width=0.9\linewidth]{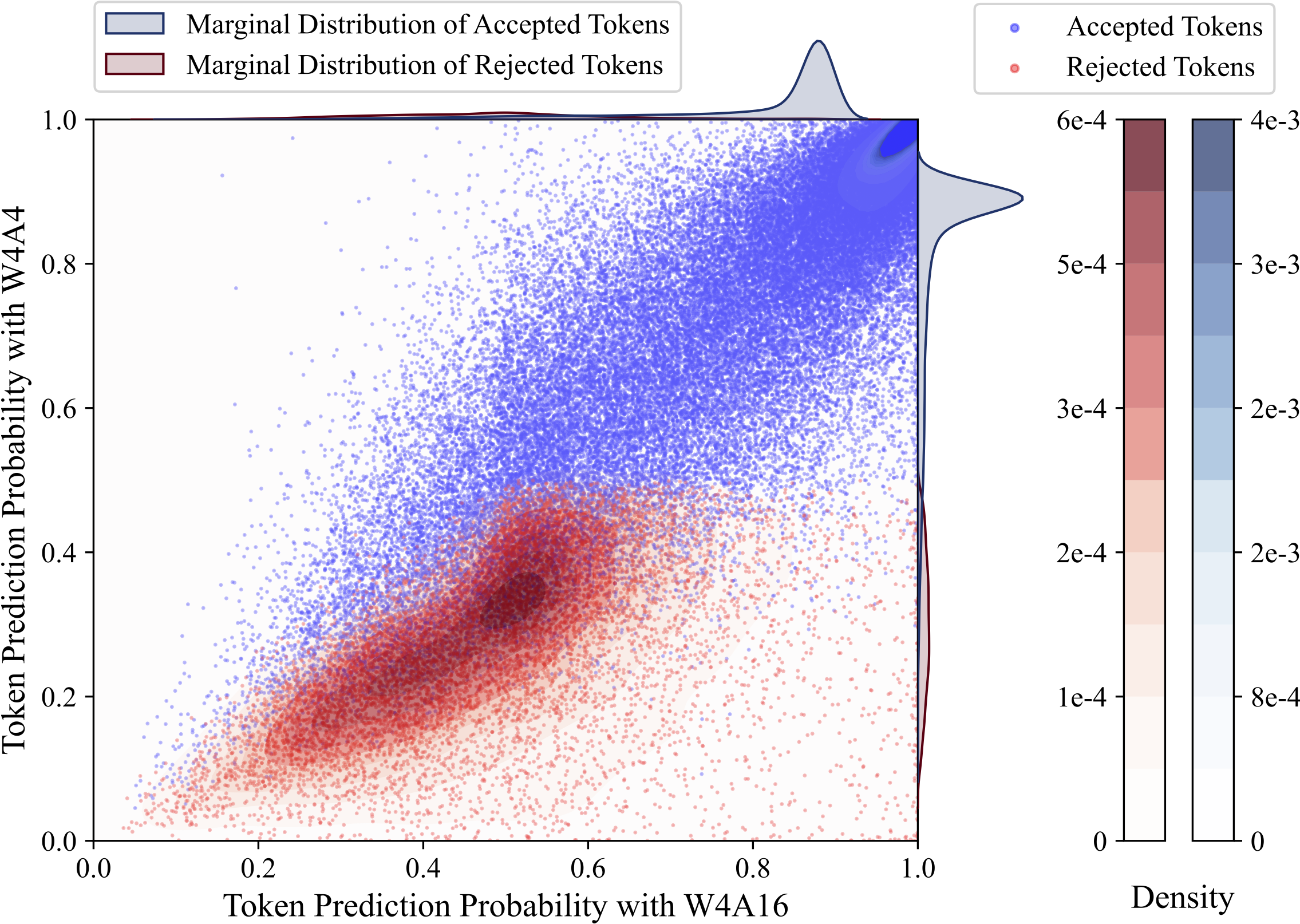}
    \caption{Scatter plot of token prediction probabilities for Atom-based W4A4 and W4A16 on GSM8K test set, along with their two-dimensional and marginal probability distributions. A striking similarity between the two quantization schemes is observed, laying the foundation of \sys{}.}
    \label{fig: similar tokens}
    \vspace{-4mm}
\end{figure}

Despite the notable performance decline caused by activation quantization, we observe, more microscopically, high similarity in top-1 token predictions between quantization schemes with high and low precision activations.
Specifically, we first employ Atom-based W4A16 greedy sampling to generate the golden token sequences for the GSM8K test set, obtaining the prediction probabilities for each top-1 answer token. Subsequently, we perform one Atom-based W4A4 forward pass (\textit{i.e.}, prefill) on the concatenated input of each question and its corresponding golden answer to acquire the token probabilities as well. This allows us to assess the prediction discrepancy between W4A4 and W4A16.
As illustrated in Figure~\ref{fig: similar tokens},  we observe that
(1)~the majority of token prediction probabilities of both W4A4 and W4A16 exceed 80\%, and most of the tokens associated with high probabilities are accepted.
(2)~Compared to accepted tokens, the number of rejected ones is negligible, underscoring the high similarity between the two quantization methods.

Combined with the analysis in Sec.~\ref{sec:reason_better}, this can be interpreted that a small set of salient token variations can trigger a snowball effect of errors, especially on multi-step reasoning tasks where the subsequent steps are closely conditioned on the previous ones, akin to findings in \cite{HalluSnowball}, thus impairing the performance of the low-precision activation scheme. 
Prior studies indicate that low similarity leads to frequent token rejections, thereby diminishing the efficiency of speculative decoding~\citep{leviathan2023fastinferencetransformersspeculative}.
The high token-level similarity we observe implies that generating high-quality outputs may only require detecting and correcting a limited number of activation quantization-induced errors. This insight motivates our proposal of a quantization-aware speculative decoding framework that leverages token generation similarity. 

\section{Method} \label{sec: method} 

\begin{figure}[t]
    \centering
    \includegraphics[width=1\linewidth]{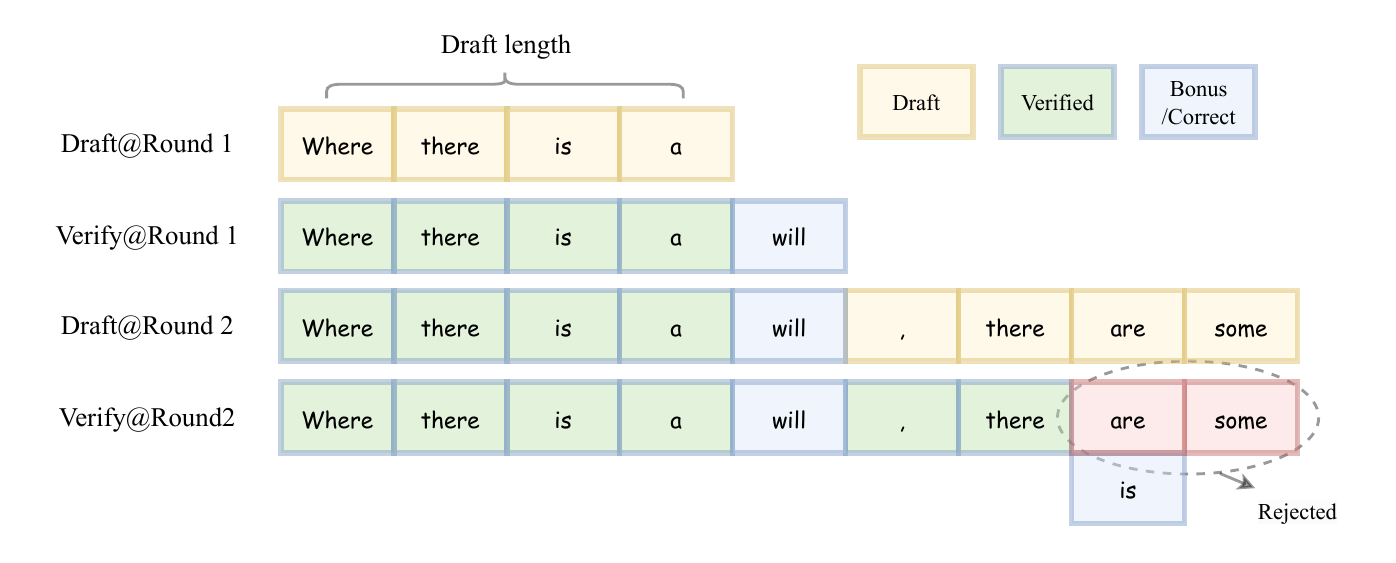}
    \vspace{-12pt}
    \caption{A mini-sample of \sys{}, where yellow, red, and blue tokens represent W4A4 draft tokens, rejected tokens, and tokens generated directly by W4A16, respectively. While these green ones  are draft tokens that have been verified and accepted by W4A16.}
    \vspace{-4mm}
    \label{fig:mini_sample}
\end{figure}

Targeting an efficient quantization scheme without sacrificing performance or increasing memory consumption, we propose a new quantization paradigm called \textbf{Spec}ulative decoding with complementary \textbf{Q}uantization execution (\sys{}). 
As shown in Figure~\ref{fig:qspec_design} and Figure~\ref{fig:mini_sample}, \sys{} employs a draft-verify pipeline for next-token prediction with varying activation precisions and shared low-precision quantized weights, instead of a single quantization scheme. Only quantization schemes are switched, and no additional weights are incorporated in this process.

\subsection{\sys{}}

\paragraph{Draft Phase.}
Current LLMs typically utilize an autoregressive process for next-token prediction. A new token is drawn from a probability distribution conditioned on all previously generated tokens. This process can be formulated as:
\begin{equation}
t_{i+1} \sim p_{i+1}(t) \coloneqq \mathcal{M}(t_{i+1} | T_{\leq i}),
\end{equation}
where $\mathcal{M}$ denotes the model including the weight and activation configurations, 
while $t_{i+1}$ and $T_{\leq i}$ represent the next predicted token and the preceding token sequence $(t_0, t_1, \ldots, t_i)$, respectively.


Compared with previous research~\citep{leviathan2023fastinferencetransformersspeculative,googlespec}, we employ a weight-shared quantization scheme with low-precision activations, rather than a standalone small-sized model, to speculate the next $\gamma$ tokens $\hat{T}_{i+1:i+\gamma}$ and their associated distributions $\hat{p}_{i+1:i+\gamma}(t)$. In $\hat{T}_{i+1:i+\gamma}$, each token $\hat{t}_j$ is sampled from $ \mathcal{M}_l(\hat{t}_{j} | T_{\leq i},\hat{T}_{i+1:j-1})$, where  \( j \in [i+1, i+\gamma] \) and  \( \mathcal{M}_l \) represents our quantized model executed with low-precision activation. Thanks to the reduced activation precision, this scheme enables fast token generation. 

\paragraph{Verify Phase.} 
To compensate for the performance decline incurred by excessive quantization, we employ a high-precision weight-only quantization scheme to verify the proposed draft token sequence. This ensures that the final generation quality aligns with that of a high-precision activation quantization scheme. All drafted tokens are verified in parallel for higher efficiency. 

Formally, the high-precision quantization scheme \( \mathcal{M}_h \) receives as input the concatenation of \( T_{\leq i} \) and \( \hat{T}_{i+1:i+\gamma} \), producing high-quality prediction probabilities \( {p}_{i+1:i+\gamma+1}(t) \) through a single forward pass. Following this, an acceptance policy \( \mathcal{A} \), which will be detailed later, is applied to rectify each drafted token sequentially. 
Once a token \( \hat{t}_{i+j} \) is rejected, all subsequent tokens are discarded, and token \( {t}_{i+j} \) is resampled according to the distribution \( p_{i+j}(t) \).
In the optimal scenario, all drafted tokens from the low-precision quantized model are accepted by the high-precision model. Subsequently, an additional token \( t_{i+\gamma+1} \) is sampled from \( p_{i+\gamma+1}(t) \). From this point, a new draft-verify cycle commences, persisting until the sequence is finalized.


\paragraph{Acceptance Policy.}
To maintain high reproducibility, both low-precision and high-precision activation quantization schemes employ greedy decoding. This means that one drafted tokens \( \hat{t}_{i+j} \) is accepted as  \({t}_{i+j} \) only when the top-1 tokens from \( {p}_{i+j} \) and \( \hat{p}_{i+j} \) coincide; otherwise, this token is rejected. 
Nonetheless, we claim that alternative strategies, as outlined in \cite{leviathan2023fastinferencetransformersspeculative}, can be directly applied to our method due to the similarities in the framework. 
Figure~\ref{fig:mini_sample} illustrates a mini-sample of this cycle with the draft token length $\gamma=4$. The model initially speculates four tokens using the W4A4 scheme. Subsequently, adhering to a predefined acceptance policy, it accepts all drafted tokens after verifying them through the W4A16 scheme. In the second loop, however, only the first two tokens are accepted. A new token ``is'' is directly derived from the prediction probability of the W4A16 scheme, and another draft-verify cycle will commence from the ninth token.

\paragraph{KV Cache Overwriting.}

A key advantage of \sys{} lies in its shared-weight architecture, which naturally aligns the behavior of low- and high-precision activations. This allows the high-precision verification stage to produce activation patterns and KV cache values that serve as high-fidelity substitutes for those generated during the low-precision draft stage. Leveraging this alignment, \sys{} overwrites the lower-quality KV caches from W4A4 with accurate A16 caches from W4A16 for accepted tokens, enabling subsequent decoding to benefit from higher-quality context. This design not only boosts token acceptance rates but also sets \sys{} apart from prior speculative decoding methods, which use separate models and cannot reuse KV representations. By sharing weights and reusing KV caches within a single model, \sys{} eliminates the need for dual cache maintenance, reducing memory usage without sacrificing accuracy.


\begin{table}[!t]
\caption{Comparison of individual quantization schemes, regular speculative decoding, and \sys{} across memory, computation, and generation aspects.}
\vspace{-1mm}
\resizebox{\columnwidth}{!}{%
\begin{tabular}{lcccccc}
\toprule
\multirow{2}{*}{Method}    & \multicolumn{2}{c}{Memory}     & \multicolumn{2}{c}{Computation}             & \multicolumn{2}{c}{Generation} \\
\cmidrule(lr){2-3} \cmidrule(lr){4-5} \cmidrule(lr){6-7}
                         & Draft Weight & Draft KV & W4A4 Kernel & Draft-Verify & High Acceptance Rate & High Fidelity \\
\midrule 
W4A16 &    \xmark  &    \xmark      &   \xmark &    \xmark                  &      -         & \cmark \\
W4A4 &    \xmark  &    \xmark      &   \cmark &    \xmark                  &      -         & \xmark \\
Speculative Decoding &    \cmark &  \cmark      &    \qmark &    \cmark                  &      \qmark        & \cmark \\
QSpec (no-overwrite)     &     \xmark~\scriptsize{(1x)}  &    \cmark~\scriptsize{(1.25x)}   &    \cmark        &    \cmark                  &      \xmark~\scriptsize{(0.8x)}    & \cmark \\
\sys{}                   &     \xmark~\scriptsize{(1x)}  &    \xmark~\scriptsize{(1x)}      &    \cmark        &    \cmark                  &      \cmark~\scriptsize{(1x)}      & \cmark \\
\bottomrule
\end{tabular}%
}
\label{tb:cost-cmp}
\end{table}

As shown in Table~\ref{tb:cost-cmp}, we compare \sys{} against individual quantization configurations (\textit{i.e.}, W4A4 and W4A16) and speculative decoding in terms of memory, computation, and generation. \sys{} provides four key benefits:
\textbf{(1) Memory-Efficient.} By sharing weights and overwriting KV caches, \sys{} incurs memory costs on par with standalone high-precision activation quantization without any memory overhead caused by speculative decoding.
\textbf{(2) No Efficiency–Efficacy Trade-off.} \sys{} leverages speculative decoding to boost efficiency while preserving output quality, avoiding the compromises in conventional quantization.
\textbf{(3) Plug-and-Play Compatibility.} \sys{} requires only an acceptance policy and a cache-overwriting step, requiring no additional training or classifiers. Hence, it can be quickly integrated into existing quantization models. 
\textbf{(4) High Acceptance Rate.} Common weights and KV cache overwriting ensure consistent predictions, leading to a high rate of token acceptance.

\subsection{Advantages of High Acceptance Rate}~\label{sec:advantages}
A key advantage of \sys{} lies in its superior acceptance rate. While state-of-the-art speculative decoding approaches such as EAGLE~\cite{eagle} and Medusa~\cite{Medusa} rely on distilled draft models and tree-structured drafting to balance speed and accuracy, \sys{} achieves high performance without such compromises. We analyze how \sys{} demonstrates superior performance compared to these methods in quantized settings below.

\noindent
\textbf{Expected average accepted token number.}
Let $p_a(t)$ be the probability of accepting a token $t$. The average accepted token number is given by Equation \ref{eq:acc_rate_tree}, where $k$ controls the tree's branching factor (width) . Larger $k$ indicates larger expected token acceptance. When $k=1$, this method reduces to standard speculative decoding.

{
\vspace{-3mm}
\small
\begin{equation}\label{eq:acc_rate_tree}
    H(k) := \sum_{l}^{\gamma} \sum_{i=1}^{k^l} \prod_{j \in Path(r, t_i)} p_a(a_j).
\end{equation}
}

\noindent
\textbf{Cost Analysis.}
Denote by $C_d(\cdot)$ and $C_p(\cdot)$ the computation cost for draft and verify, The first argument indicates the number of sequences in a draft tree, and the second argument is the prefix sequence length. For a prefix with length $s$, we have \textit{per-valid-token} cost $v$ by dividing the total cost by the number of accepted tokens in Equation:

{
\vspace{-3mm}
\small
\begin{equation}\label{eq:per_cost}
\resizebox{0.9\columnwidth}{!}{%
$\displaystyle
\begin{aligned}
v
&= 
\frac{
    \overbrace{C_d(1; s) 
        + \cdots 
        + C_d\bigl(k^{\gamma{}-1}; s + \gamma{} - 1\bigr)}^{\text{Drafting cost}}
    \;+\;
    \overbrace{C_p\bigl(k^{\gamma{}-1}; s+\gamma{}\bigr)}^{\text{Verify cost}}
}{
    \underbrace{H(k)}_{\text{Accepted tokens}}
}
\sim 
\frac{C(k)}{H(k)}.
\end{aligned}
$%
}
\end{equation}
}

\noindent
\textbf{Batch serving efficiency of \sys{}.} The efficacy of speculative decoding relies on the assumption that verification cost \( C_p \) approximates the target model’s decoding cost \( C_d \), leveraging underutilized resources for parallel verification.\citep{googlespec,leviathan2023fastinferencetransformersspeculative,yan2025decodingspeculativedecoding} This holds in single-request, memory-bound scenarios. However, tree-structured drafting, with verification cost \( C_p(k^{\gamma-1}; s + \gamma) \), scales poorly as \( k^{\gamma-1} \) far exceeds non-tree sequence length \( \gamma \). Therefore, in batched serving, tree-based \( C_p \) enters compute-bound territory easily, significantly surpassing \( C_d \) and incurring high memory overhead, invalidating the assumption. \sys{}, using W4A4 quantization, avoids tree structures, maintaining low \( C_p \) and memory usage and balancing the loads between drafting and verification stages, \sys{} delays the onset of this inefficiency. In Section~\ref{sec: exp}, \sys{} demonstrates excellent speedup even at a batch size of 32, also outperforming EAGLE~\citep{eagle} in comparison experiments.

\section{Experiments} \label{sec: exp}  
Our evaluation answers three key questions:
\begin{enumerate}[label=Q\arabic*:, itemsep=0pt, topsep=0pt, partopsep=0pt]
    \item Does \sys{} preserve the quality of high-precision weight-only quantization? (Sec.~\ref{sec: acc_res})
    \item Does \sys{} speed up high-precision weight-only quantization methods, and surpass speculative decoding methods in quantized scenarios? (Sec.\ref{sec: perf_ana})
    \item How do various factors (\textit{e.g.}, quantization method, sequence length) influence the acceptance rate and acceleration performance of \sys{}? (Sec.~\ref{sec: ablation})
\end{enumerate}


\subsection{General Setup} \label{sec:setup}
\textbf{Benchmarks.}
We assess \sys{} with two primary criteria: (1) generation fidelity and (2) end-to-end serving speedup. For fidelity evaluation, we adopt not only traditional tasks, including PIQA (500, 10-shot)~\citep{piqa}, WinoGrande (500, 5-shot)~\citep{ai2:winogrande}, and WikiText2~\citep{wikitext2}, but also challenging multi-step reasoning tasks such as GSM8K (All, 8-shot)~\citep{gsm8k}, MATH (All, 4-shot)~\citep{MATH}, MBPP (200, 0-shot)~\citep{MBPP}, and HumanEval (All, 0-shot)~\citep{HUMAN_EVAL}. To measure the acceleration, we use all the above reasoning tasks and two additional chatbot datasets, namely ShareGPT~\citep{sharegpt52k} and LMsys-1K~\citep{lmsys}. Due to space constraints, we present results for GSM8K, HumanEval, and LMsys-1K in the main text, with remaining results detailed in the Appendix~\ref{sec:app_full_res}. Following the setup of Atom~\citep{Atom}, we randomly sampled the dataset for the request prompts to reduce the workload. Due to memory limitations, we vary the batch size from 8 to 32 and serve all requests in a first-come, first-served (FCFS) manner. Once any request is finished, we refill the batch, adhering to the continuous batching approach of ORCA~\citep{orca}. All experiments use greedy sampling for token generation.

\textbf{Models.}
To assess the effectiveness and scalability of our approach, we conduct experiments using multiple models from the Llama family~\citep{llama2, llama3} 
with varying scales and capacities:
Llama3.2-3b, Llama2-7b, Llama3-8b-instruct, and Llama2-13b.

\begin{table*}[t]
\centering
\caption{Performance of different quantization methods across multiple general and reasoning benchmarks: PIQA, WinoGrande, GSM8K, MATH, MBPP, and HumanEval.
The quality degradation ratio is calculated by $ \frac{W4A4}{W4A16} - 1$. }
\resizebox{\linewidth}{!}{%
\begin{tabular}{@{}ccccccccc@{}}
\toprule
\multirow{2}{*}{\textbf{Method}} & \multirow{2}{*}{\textbf{Quantization}} & \textbf{WikiText-2}
& \textbf{PIQA} & \textbf{WinoGrande} & \textbf{GSM8K} & \textbf{MATH} & \textbf{MBPP} & \textbf{HumanEval} \\
& & PPL $\downarrow$ & EM (\%) $\uparrow$ & EM (\%) $\uparrow$ & EM (\%) $\uparrow$ & EM (\%) $\uparrow$ & Pass@1 (\%) $\uparrow$ & Pass@1 (\%) $\uparrow$ \\
\midrule
\multirow{3}{*}{Atom} 
 & W16A16 & 7.73 & 76.8 & 61.4 & 76.2 & 24.9 & 42.5 & 53.0 \\
 & W4A16 & 7.87 & 74.8 & 62.0 & 73.4 & 24.3 & 42.0 & 52.4 \\
 & \sys{} & 7.87 & 75.0 & 62.0 & 73.4 & 24.3 & 40.5 & 52.4 \\
& W4A4 & 8.6 (+9.58\%) & 65.8 (-12.03\%) & 56.2 (-9.35\%) & 54.7 (-25.47\%) & 15.5 (-36.21\%) & 33.0 (-21.43\%) & 31.7 (-39.50\%) \\
\midrule
\multirow{3}{*}{QuaRot} 
 & W16A16 & 7.73 & 76.8 & 61.4 & 76.2 & 24.9 & 42.5 & 53.0 \\
 & W4A16 & 8.58 & 74.2 & 59.4 & 70.5 & 24.7 & 40.0 & 45.7 \\
 & \sys{} & 8.58 & 74.4 & 59.2 & 71.0 & 24.7 & 40.5 & 47.6 \\
& W4A4 & 10.2 (+19.24\%) & 62.6 (-15.63\%) & 53.8 (-9.43\%) & 42.0 (-40.43\%) & 12.3 (-51.11\%) & 28.5 (-28.75\%) & 28.0 (-38.73\%) \\
\bottomrule
\end{tabular}%
}
\label{tab:accuracy_main}
\end{table*}

\textbf{Implementations.}
All experiments are performed on a node equipped with four NVIDIA L20 GPUs (48GB HBM each) running CUDA 12.5.
To demonstrate the versatility of \sys{}, we implement two SOTA 4-bit quantization methods, namely Atom~\citep{Atom} and QuaRot~\citep{quarot}. For W4A16 configurations, we incorporate AWQ-style~\citep{awq} weight dequantization logic for runtime inference. We select Atom to showcase the acceleration of \sys{}. We use these Group-wise quantization schemes with a group size of 128. We configure the default draft token length $\gamma$ to 3. The implementation of \sys{} follows Atom's setup, incorporating flashinfer~\citep{flashinfer}.  

\textbf{Baselines.}
We evaluate \sys{} against baseline quantization configurations: W4A4, W4A16, and W16A16. We also include EAGLE~\citep{eagle}, which employs a pruned draft model with tree-structured speculative decoding, as a baseline to compare with regular speculative decoding. In our quantized experiments, we utilize FP16 precision for the EAGLE draft model and EAGLE-Quant (W4A16) for the target model. This choice was necessitated by two factors: (1) the official EAGLE quantization implementation (fast-gpt) lacks efficient batching support, and (2) applying GPTQ quantization to the EAGLE draft model resulted in substantial degradation of the acceptance rate.



\subsection{Fidelity Evaluation} \label{sec: acc_res}

\textbf{\sys{} effectively maintains the generation quality of W4A16, whereas W4A4 does not.} As listed in Table~\ref{tab:accuracy_main},
with the draft verification of W4A16, \sys{} exhibits only minimal performance fluctuations compared to W4A16. This negligible variation may stem from the non-deterministic algorithms of PyTorch~\citep{pytorch2024reproducibility} or occasional cases where two tokens have the same maximum prediction probability.
In contrast, W4A4 experiences a substantial performance decline exceeding 10\% across most tasks, with the reduction becoming more pronounced as task difficulty increases. For instance, compared to GSM8K and MBPP, the performance drop for W4A4 is much greater on the more challenging MATH and HumanEval tasks, showing declines of 51.11\% and 38.73\%, respectively. 
On the other hand, this also highlights the higher sensitivity of multi-step reasoning tasks to the negative effects of quantization compared to regular tasks, such as WikiText-2
and WinoGrande.
This observation fully aligns with our earlier analysis in Sec.~\ref{sec: motivation}, encouraging the incorporation of multi-step reasoning tasks into quantization evaluation. 


\begin{table}[!t]
    \vspace{-2mm}
    \centering
    \caption{%
        Comparison of token generation throughput across different model sizes, 
        quantization configurations, and batch sizes for various datasets. 
        All values are measured in token/s.
        "Avg." denotes the average speedup ratio for the corresponding row or column.
    }
    \label{tab:performance_comparison_l20}
    \scriptsize
    \setlength{\tabcolsep}{2pt}
    \renewcommand{\arraystretch}{0.9}
    \begin{tabular}{C{1.0cm} C{1.1cm} C{0.9cm} C{1.2cm} C{1.2cm} C{1.2cm}}
        \toprule
        \multirow{2}{*}{\makecell{Model}} & \multirow{2}{*}{\makecell{Method}} & \multirow{2}{*}{\makecell{Batch}} & \multirow{2}{*}{\makecell{GSM8K}} & \multirow{2}{*}{\makecell{HumanEval}} & \multirow{2}{*}{\makecell{LMsys-1k}} \\
        & & & & & \\
        \midrule
        \multirow{14}{*}{\textbf{3B}}%
        & \multirow{3}{*}{W4A4}%
        & 8 & 804.7 & 892.6 & 990.3 \\
        & & 16 & 1109.1 & 1289.8 & 1581.0 \\
        & & 32 & 1424.3 & 1488.2 & 2194.4 \\
        \cmidrule(lr){2-6}
        & \multirow{3}{*}{W4A16}%
        & 8 & 420.0 & 535.7 & 559.8 \\
        & & 16 & 578.5 & 804.4 & 925.8 \\
        & & 32 & 726.3 & 954.4 & 1336.4 \\
        \cmidrule(lr){2-6}
        & \multirow{5}{*}{\sys{}}%
        & 8 & 594.1 (1.41$\times$) & 723.6 (1.35$\times$) & 738.8 (1.32$\times$) \\
        & & 16 & 811.5 (1.40$\times$) & 1042.1 (1.30$\times$) & 1171.4 (1.27$\times$) \\
        & & 32 & 1030.4 (1.42$\times$) & 1248.5 (1.31$\times$) & 1576.0 (1.18$\times$) \\
        \cmidrule(lr){3-6}
        \rowcolor{gray!15}%
        & & Avg. & 1.41$\times$ & 1.32$\times$ & 1.25$\times$ \\
        \midrule
        \multirow{14}{*}{\textbf{7B}}%
        & \multirow{3}{*}{W4A4}%
        & 8 & 349.5 & 471.2 & 419.4 \\
        & & 16 & 496.6 & 749.5 & 642.6 \\
        & & 32 & 620.0 & 1043.9 & 865.5 \\
        \cmidrule(lr){2-6}
        & \multirow{3}{*}{W4A16}%
        & 8 & 165.0 & 240.2 & 220.2 \\
        & & 16 & 231.8 & 407.3 & 358.0 \\
        & & 32 & 268.9 & 555.9 & 470.1 \\
        \cmidrule(lr){2-6}
        & \multirow{5}{*}{\sys{}}%
        & 8 & 253.7 (1.54$\times$) & 350.9 (1.46$\times$) & 310.3 (1.41$\times$) \\
        & & 16 & 359.8 (1.55$\times$) & 555.2 (1.36$\times$) & 473.1 (1.32$\times$) \\
        & & 32 & 441.8 (1.64$\times$) & 749.4 (1.35$\times$) & 628.4 (1.34$\times$) \\
        \cmidrule(lr){3-6}
        \rowcolor{gray!15}%
        & & Avg. & 1.58$\times$ & 1.39$\times$ & 1.36$\times$ \\
        \midrule
        \multirow{14}{*}{\textbf{13B}}%
        & \multirow{3}{*}{W4A4}%
        & 8 & 194.7 & 261.5 & 228.2 \\
        & & 16 & 288.3 & 424.9 & 348.4 \\
        & & 32 & 369.8 & 665.4 & 508.8 \\
        \cmidrule(lr){2-6}
        & \multirow{3}{*}{W4A16}%
        & 8 & 94.8 & 140.0 & 127.9 \\
        & & 16 & 136.1 & 236.9 & 207.2 \\
        & & 32 & 207.5 & 365.5 & 287.4 \\
        \cmidrule(lr){2-6}
        & \multirow{5}{*}{\sys{}}%
        & 8 & 148.2 (1.56$\times$) & 201.2 (1.44$\times$) & 174.0 (1.36$\times$) \\
        & & 16 & 212.8 (1.56$\times$) & 323.3 (1.36$\times$) & 266.9 (1.29$\times$) \\
        & & 32 & 266.6 (1.28$\times$) & 483.0 (1.32$\times$) & 379.3 (1.32$\times$) \\
        \cmidrule(lr){3-6}
        \rowcolor{gray!15}%
        & & Avg. & 1.47$\times$ & 1.37$\times$ & 1.32$\times$ \\
        \bottomrule
    \end{tabular}
    \vspace{-2mm}
\end{table}

\subsection{Acceleration Evaluation} \label{sec: perf_ana}

\textbf{\sys{} exhibits a substantial efficiency boost compared to W4A16.}
In Table~\ref{tab:performance_comparison_l20}, we present the token generation throughput for both \sys{} and W4A16 across different model sizes, quantization configurations, and batch sizes on diverse datasets. On average, \sys{} achieves a throughput increase of $1.38\times$ over W4A16 across all settings, with a peak improvement of $1.64\times$.



\textbf{Larger models tend to yield better speedup ratios.}
We observe a consistent acceleration trend as the base model scales, demonstrating the promising scalability of our approach with larger models. While further validation is needed, resource constraints necessitate addressing this in future work.

\begin{figure}[t]
    \centering
    \includegraphics[width=\linewidth]{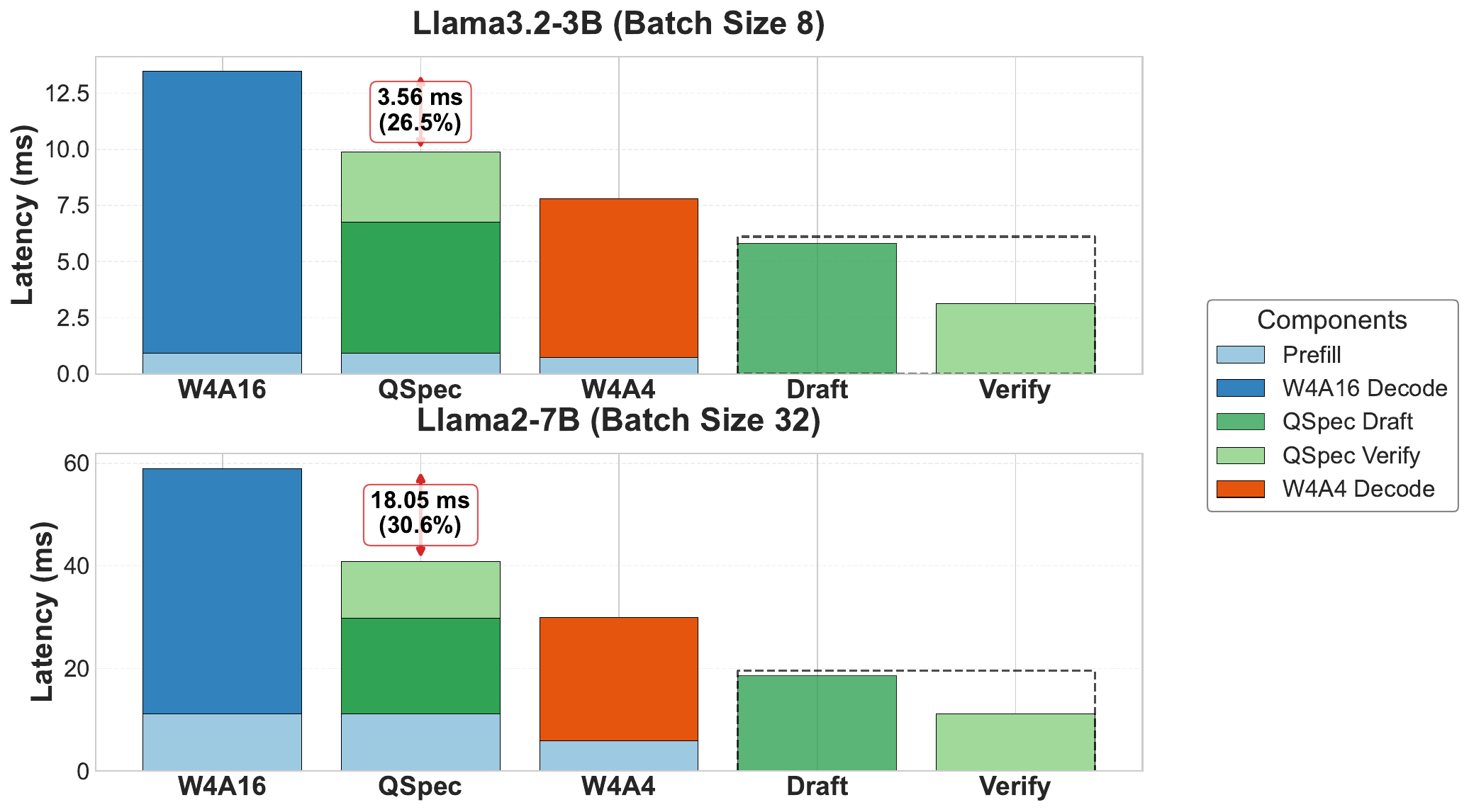}
    \vspace{-5mm}
    \caption{Per-valid-token latency decomposition for different methods. The latency of \sys{} is further decomposed into draft and verify categories for details.}
    \label{fig:per_token_analysis}
    \vspace{-12pt}
\end{figure}

\textbf{\sys{} reduces latency through fast drafting and parallel verifying.} As illustrated in Figure~\ref{fig:per_token_analysis}, we compute the per-valid-token latency by dividing the total latency by only the number of accepted tokens before averaging on all evaluation datasets. 
Notably, \sys{} achieves remarkable latency savings ranging from 26.5\% to 30.6\%. 
Besides, the per-token latency is further decomposed into two parts: draft and verify.
Clearly, the primary gains of \sys{} arise from the rapid drafting capability and the reduced latency achieved through the parallel verification of multiple tokens.

\begin{table}[!t]
    \centering
    \vspace{-1.5mm}
    \caption{Performance comparison of EAGLE-Quant, \sys, W4A16, and W4A4 on Llama-2-7b-chat-hf across different batch sizes and benchmarks. “OOM” indicates out-of-memory. Better cases for \sys{} or EAGLE is marked in gray. In the case of batch size=8, the speedup ratio of \sys{} compared to EAGLE is indicated in parentheses next to the data points.}
    \label{tab:quant_comparison}
    \vspace{-3mm}
    \scriptsize
    \setlength{\tabcolsep}{3pt}
    \renewcommand{\arraystretch}{1.0}
    \begin{tabular}{C{1.2cm} C{1.0cm} C{1.2cm} C{1.3cm} C{1.3cm}}
        \toprule
        \multirow{2}{*}{\makecell{Method}} & \multirow{2}{*}{\makecell{Batch \\ Size}} & \multirow{2}{*}{\makecell{GSM8K \\ (8-shot)}} & \multirow{2}{*}{\makecell{HumanEval \\ (0-shot)}} & \multirow{2}{*}{\makecell{LMsys-1k}} \\
        & & & & \\
        \midrule
        \rowcolor{gray!15}
        \multirow{3}{*}{EAGLE}
        & 1  & 65.81  & 49.15  & 71.29 \\
        & 8  & 140.16 & 136.86 & 167.57 \\
        & 16 & OOM    & OOM    & OOM \\
        \midrule
        & 1  & 51.25  & 54.22  & 56.14 \\
        \rowcolor{gray!15}
        \sys{} & 8  & 208.95 (1.49$\times$) & 185.99 (1.36$\times$) & 260.48 (1.55$\times$) \\
        \rowcolor{gray!15}
        & 16 & 292.82 & 255.11 & 463.35 \\
        \midrule
        \multirow{3}{*}{W4A16}
        & 1  & 59.80  & 72.04  & 72.27 \\
        & 8  & 146.34 & 163.54 & 213.66 \\
        & 16 & 190.09 & 211.49 & 371.38 \\
        \midrule
        \multirow{3}{*}{W4A4}
        & 1  & 64.79  & 73.47  & 71.55 \\
        & 8  & 284.84 & 256.54 & 393.12 \\
        & 16 & 401.77 & 330.71 & 713.67 \\
        \bottomrule
    \end{tabular}
    \vspace{-3mm}
\end{table}

\textbf{\sys{} offers better memory and batching efficiency than prior speculative decoding methods.} Table~\ref{tab:quant_comparison} compares \sys{} and EAGLE on Llama-2-7b-chat-hf. EAGLE delivers optimal performance for single-sequence inputs (batch size=1). However, its efficiency degrades as the batch size grows (8 and 16). This observation aligns with our earlier analysis in Sec.~\ref{sec:advantages}: EAGLE’s tree-structured drafting mechanism, designed to reconcile discrepancies between the draft and target models, introduces additional latency, and reduces the gains from higher acceptance rates in batched serving under quantization. Furthermore, the increasing key-value (KV) storage of EAGLE’s draft model leads to out-of-memory (OOM) issues at batch size 16. In contrast, \sys{} demonstrates superior scalability and memory efficiency.

\textbf{\sys{} excels in real-world deployment.} \label{sec: exp_vllm} We integrated \sys{} into vLLM\citep{EfficientMemoryManagement_Kwon2023} to validate its performance in real-world serving scenarios. Despite a suboptimal implementation, our experiments demonstrate an average speedup of 1.24$\times$, with effective acceleration even at a batch size of 32. 
Details are provided in Appendix~\ref{app:vLLM}.

\subsection{Ablation Study}~\label{sec: ablation}

\begin{figure}[t]
    \centering
    \includegraphics[width=\linewidth]{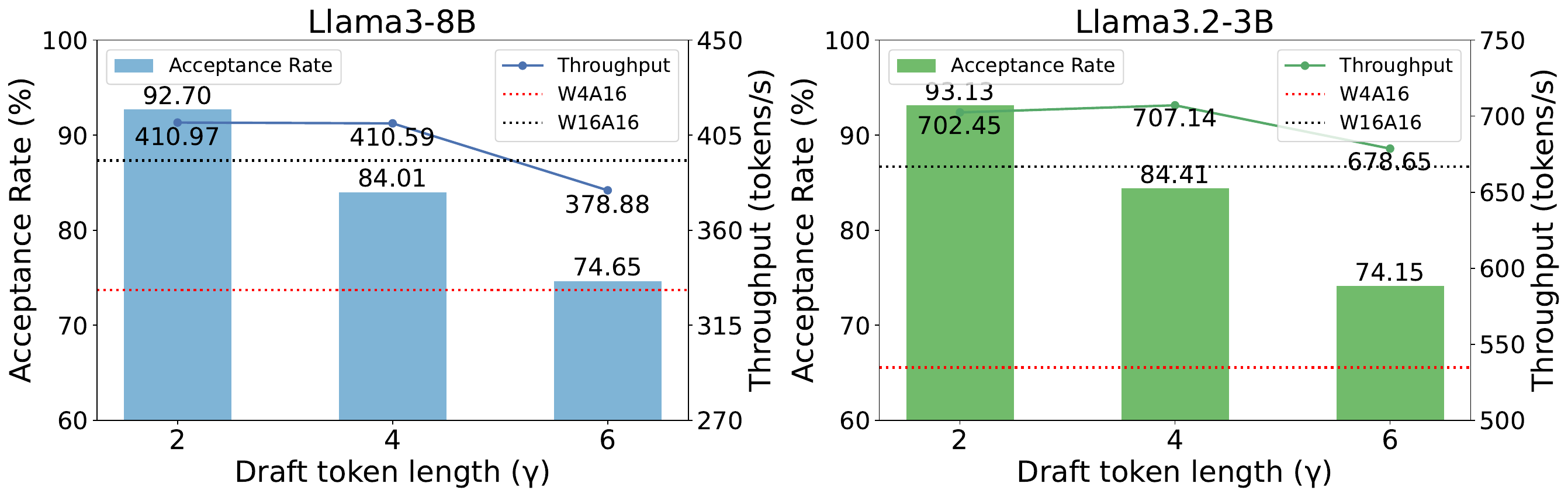}
    \vspace{-5mm}
    \caption{Acceptance rate and throughput of Llama3.2-3b (batch size 8) and Llama3-8b-instruct (batch size 16) with respect to the draft token length $\gamma$.}
    \label{fig:draft token length}
    \vspace{-3mm}
\end{figure}

\textbf{Ablation on draft token length.}
To assess parameter sensitivity, we vary the draft token lengths $\gamma$, the only hyper-parameter of \sys{}, from 2 to 6 across all the benchmarks using Llama3.2-3b and Llama3-8b-instruct models.
As depicted in Figure~\ref{fig:draft token length},
an increase in \(\gamma\) leads to a gradual decline in the token acceptance rate, since all subsequent tokens are discarded once a token is rejected.
Nevertheless, even at \(\gamma = 6\), the token acceptance rate remains relatively high, approximately $74\%$, compared to $28\sim 58\%$ in 160m-7b draft-target model pair under $\gamma=5$ in conventional speculative decoding~\citep{liu2024onlinespeculativedecoding}.
Additionally, a consistent improvement in throughput is observed compared to W4A16, indicating the robustness of \sys{} with respect to \(\gamma\). More comprehensive comparison is provided in Appendix~\ref{sec:app_comphre}.

\textbf{Ablation on base quantization method.} As shown in Appendix Table~\ref{tab:acceptance_rate_for_different_quant_methods}, \sys{} consistently achieves high acceptance rates across diverse quantization methods and datasets.

\section{Related Work} \label{sec: related}
\vspace{-2mm}

This work builds on two lines of research: quantization and speculative decoding. Weight-only quantization (W4A16) offers better accuracy, while joint weight-activation methods (W4A4) enable faster inference but degrade performance on complex tasks.
Speculative decoding improves efficiency by verifying drafted tokens, but existing approaches typically require retraining and are less effective under quantization. We provide detailed comparisons in Appendix~\ref{sec:appendix_related}.

\section{Conclusion} \label{sec:conclusion}
\vspace{-2mm}

We begin by showing that multi-step reasoning tasks reveal performance degradation from activation quantization more clearly than current evaluation protocols, and encourage their incorporation for more comprehensive assessment. Leveraging high token-level similarities, we propose \sys{}, a novel quantization paradigm that seamlessly combines two complementary weight-shared quantization schemes through speculative decoding.
Empirically, \sys{} delivers significant acceleration—up to $1.64\times$—without quality loss across diverse settings. With consistent memory usage and a plug-and-play design, \sys{} offers a practical and scalable solution for high-fidelity quantization, especially under memory constraints.

\section*{Acknowledgements}
This work was supported in part by grants from Hong Kong RGC under the contracts 17204423, 17205824, C7004-22G (CRF),  C5032-23G (CRF) and T43-513/23-N (TRS).

\section{Limitations and Impacts}  \label{sec:impact}

\subsection{Impact Statement}

This paper introduces \sys{}, a novel quantization paradigm that synergies complementary quantization schemes through speculative decoding to enhance computational efficiency while preserving model fidelity. The impact of \sys{} is twofold.  

From an academic perspective, \sys{} establishes a new paradigm that decouples efficiency from quality preservation—a longstanding trade-off in prior quantization research. 
This is achieved by the complementary quantization schemes:
a low-precision activation-weight joint quantization for fast token drafting, and a high-precision weight-only quantization for accurate verification, enabling the independent optimization for efficiency and quality.
This illuminates the pursuit of extreme efficiency in quantization schemes (\textit{e.g.}, W4A4) without the concern of performance degradation.

For industry applications, \sys{} provides a practical solution to accelerate inference without compromising output quality through efficient low-precision kernels. This prompts hardware vendors to reconsider their architectural support for low-precision execution, including specialized instruction set architectures (ISAs) and memory subsystems. 
Besides, the plug-and-play property of \sys{} further facilitates seamless integration into existing deployments of quantized models in memory-constraint scenarios (\textit{e.g.}, edge devices).

Our research, focused on improving the computational efficiency of language model serving systems, is not anticipated to have direct negative social impact.

\subsection{Limitation Discussion} \label{app:limit}

The superior performance of \sys{} relies on the high acceptance rate,
particularly in small to moderate batch-size scenarios where the throughput gap between low- and high-precision quantization becomes pronounced. In contrast, traditional tree-based speculative decoding methods falter in batch serving, as discussed in Sec.~\ref{sec: exp}, making \sys{}’s advantages most evident in these settings. However, \sys{} exhibits potential limitations in single-request scenarios, where other methods are preferentially optimized.
With the popularity of LLMs and increasing batch serving, regular speculative decoding methods, including Medusa and EAGLE, degrade significantly (\textit{e.g.}, Medusa’s speedup drops below 1 at batch size of 16 in their Figure 22~\cite{Medusa}), whereas \sys{} excels. 




To further address the limitations of \sys{} in single-request scenarios, future research will focus on leveraging its high acceptance rate and reducing the overhead of the draft stage. Specifically, we aim to develop adaptive mechanisms that dynamically adjust the draft model’s sparsity, balancing latency and acceptance rate to achieve robust performance across both single-request and batch-size settings. 
Additionally, exploring hardware-aware optimizations, such as tailored low-precision kernels for resource-constrained devices, will enhance \sys{}’s applicability in edge deployments.
Furthermore, integrating \sys{} into popular repositories (\textit{e.g.}, vLLM\footnote{\url{https://docs.vllm.ai/en/latest/}})
is also part of our future work.
By unifying these advancements, \sys{} can evolve into a versatile quantization framework, delivering consistent acceleration and fidelity for diverse inference scenarios, from personal devices to large-scale serving systems.

\bibliography{paper}

\begin{thebibliography}{45}
\providecommand{\natexlab}[1]{#1}

\bibitem[{Ainslie et~al.(2023)Ainslie, Lee-Thorp, de~Jong, Zemlyanskiy, Lebrón, and Sanghai}]{GQA}
Joshua Ainslie, James Lee-Thorp, Michiel de~Jong, Yury Zemlyanskiy, Federico Lebrón, and Sumit Sanghai. 2023.
\newblock \href {https://arxiv.org/abs/2305.13245} {Gqa: Training generalized multi-query transformer models from multi-head checkpoints}.
\newblock \emph{Preprint}, arXiv:2305.13245.

\bibitem[{Ashkboos et~al.(2024)Ashkboos, Mohtashami, Croci, Li, Jaggi, Alistarh, Hoefler, and Hensman}]{quarot}
Saleh Ashkboos, Amirkeivan Mohtashami, Maximilian~L. Croci, Bo~Li, Martin Jaggi, Dan Alistarh, Torsten Hoefler, and James Hensman. 2024.
\newblock \href {https://arxiv.org/abs/2404.00456} {Quarot: Outlier-free 4-bit inference in rotated llms}.
\newblock \emph{Preprint}, arXiv:2404.00456.

\bibitem[{Austin et~al.(2021)Austin, Odena, Nye, Bosma, Michalewski, Dohan, Jiang, Cai, Terry, Le, and Sutton}]{MBPP}
Jacob Austin, Augustus Odena, Maxwell Nye, Maarten Bosma, Henryk Michalewski, David Dohan, Ellen Jiang, Carrie Cai, Michael Terry, Quoc Le, and Charles Sutton. 2021.
\newblock \href {https://arxiv.org/abs/2108.07732} {Program synthesis with large language models}.
\newblock \emph{Preprint}, arXiv:2108.07732.

\bibitem[{Bisk et~al.(2020)Bisk, Zellers, Bras, Gao, and Choi}]{piqa}
Yonatan Bisk, Rowan Zellers, Ronan~Le Bras, Jianfeng Gao, and Yejin Choi. 2020.
\newblock Piqa: Reasoning about physical commonsense in natural language.
\newblock In \emph{Thirty-Fourth AAAI Conference on Artificial Intelligence}.

\bibitem[{Cai et~al.(2024)Cai, Li, Geng, Peng, Lee, Chen, and Dao}]{Medusa}
Tianle Cai, Yuhong Li, Zhengyang Geng, Hongwu Peng, Jason~D. Lee, Deming Chen, and Tri Dao. 2024.
\newblock \href {https://arxiv.org/abs/2401.10774} {Medusa: Simple llm inference acceleration framework with multiple decoding heads}.
\newblock \emph{Preprint}, arXiv:2401.10774.

\bibitem[{Chen et~al.(2023)Chen, Borgeaud, Irving, Lespiau, Sifre, and Jumper}]{googlespec}
Charlie Chen, Sebastian Borgeaud, Geoffrey Irving, Jean-Baptiste Lespiau, Laurent Sifre, and John Jumper. 2023.
\newblock Accelerating large language model decoding with speculative sampling.
\newblock \emph{arXiv preprint arXiv:2302.01318}.

\bibitem[{Chen et~al.(2021)Chen, Tworek, Jun, Yuan, de~Oliveira~Pinto, Kaplan, Edwards, Burda, Joseph, Brockman, Ray, Puri, Krueger, Petrov, Khlaaf, Sastry, Mishkin, Chan, Gray, Ryder, Pavlov, Power, Kaiser, Bavarian, Winter, Tillet, Such, Cummings, Plappert, Chantzis, Barnes, Herbert-Voss, Guss, Nichol, Paino, Tezak, Tang, Babuschkin, Balaji, Jain, Saunders, Hesse, Carr, Leike, Achiam, Misra, Morikawa, Radford, Knight, Brundage, Murati, Mayer, Welinder, McGrew, Amodei, McCandlish, Sutskever, and Zaremba}]{HUMAN_EVAL}
Mark Chen, Jerry Tworek, Heewoo Jun, Qiming Yuan, Henrique~Ponde de~Oliveira~Pinto, Jared Kaplan, Harri Edwards, Yuri Burda, Nicholas Joseph, Greg Brockman, Alex Ray, Raul Puri, Gretchen Krueger, Michael Petrov, Heidy Khlaaf, Girish Sastry, Pamela Mishkin, Brooke Chan, Scott Gray, and 39 others. 2021.
\newblock \href {https://arxiv.org/abs/2107.03374} {Evaluating large language models trained on code}.

\bibitem[{Chen et~al.(2024)Chen, May, Svirschevski, Huang, Ryabinin, Jia, and Chen}]{sequoia}
Zhuoming Chen, Avner May, Ruslan Svirschevski, Yuhsun Huang, Max Ryabinin, Zhihao Jia, and Beidi Chen. 2024.
\newblock \href {https://arxiv.org/abs/2402.12374} {Sequoia: Scalable, robust, and hardware-aware speculative decoding}.
\newblock \emph{Preprint}, arXiv:2402.12374.

\bibitem[{Clark et~al.(2018)Clark, Cowhey, Etzioni, Khot, Sabharwal, Schoenick, and Tafjord}]{allenai:arc}
Peter Clark, Isaac Cowhey, Oren Etzioni, Tushar Khot, Ashish Sabharwal, Carissa Schoenick, and Oyvind Tafjord. 2018.
\newblock Think you have solved question answering? try arc, the ai2 reasoning challenge.
\newblock \emph{arXiv:1803.05457v1}.

\bibitem[{Cobbe et~al.(2021)Cobbe, Kosaraju, Bavarian, Chen, Jun, Kaiser, Plappert, Tworek, Hilton, Nakano, Hesse, and Schulman}]{gsm8k}
Karl Cobbe, Vineet Kosaraju, Mohammad Bavarian, Mark Chen, Heewoo Jun, Lukasz Kaiser, Matthias Plappert, Jerry Tworek, Jacob Hilton, Reiichiro Nakano, Christopher Hesse, and John Schulman. 2021.
\newblock \href {https://arxiv.org/abs/2110.14168} {Training verifiers to solve math word problems}.
\newblock \emph{Preprint}, arXiv:2110.14168.

\bibitem[{Deng et~al.(2024)Deng, Zhao, Hessel, Ren, Cardie, and Choi}]{deng2024wildvisopensourcevisualizer}
Yuntian Deng, Wenting Zhao, Jack Hessel, Xiang Ren, Claire Cardie, and Yejin Choi. 2024.
\newblock \href {https://arxiv.org/abs/2409.03753} {Wildvis: Open source visualizer for million-scale chat logs in the wild}.
\newblock \emph{Preprint}, arXiv:2409.03753.

\bibitem[{Dong et~al.(2024)Dong, Cheng, Qin, and Wang}]{Dong2024QAQQA}
Shichen Dong, Wenfang Cheng, Jiayu Qin, and Wei Wang. 2024.
\newblock \href {https://api.semanticscholar.org/CorpusID:268264510} {Qaq: Quality adaptive quantization for llm kv cache}.
\newblock \emph{ArXiv}, abs/2403.04643.

\bibitem[{Dubey et~al.(2024)Dubey, Jauhri, Pandey, Kadian, Al-Dahle, Letman, Mathur, Schelten, Yang, Fan et~al.}]{llama3}
Abhimanyu Dubey, Abhinav Jauhri, Abhinav Pandey, Abhishek Kadian, Ahmad Al-Dahle, Aiesha Letman, Akhil Mathur, Alan Schelten, Amy Yang, Angela Fan, and 1 others. 2024.
\newblock The llama 3 herd of models.
\newblock \emph{arXiv preprint arXiv:2407.21783}.

\bibitem[{Elhoushi et~al.(2024)Elhoushi, Shrivastava, Liskovich, Hosmer, Wasti, Lai, Mahmoud, Acun, Agarwal, Roman, Aly, Chen, and Wu}]{layerskip}
Mostafa Elhoushi, Akshat Shrivastava, Diana Liskovich, Basil Hosmer, Bram Wasti, Liangzhen Lai, Anas Mahmoud, Bilge Acun, Saurabh Agarwal, Ahmed Roman, Ahmed Aly, Beidi Chen, and Carole-Jean Wu. 2024.
\newblock \href {https://api.semanticscholar.org/CorpusID:269362647} {Layerskip: Enabling early exit inference and self-speculative decoding}.
\newblock \emph{ArXiv}, abs/2404.16710.

\bibitem[{Guo et~al.(2024{\natexlab{a}})Guo, Zhu, Yang, Xie, Dong, Zhang, Chen, Bi, Wu, Li, Luo, Xiong, and Liang}]{ds_code}
Daya Guo, Qihao Zhu, Dejian Yang, Zhenda Xie, Kai Dong, Wentao Zhang, Guanting Chen, Xiao Bi, Y.~Wu, Y.~K. Li, Fuli Luo, Yingfei Xiong, and Wenfeng Liang. 2024{\natexlab{a}}.
\newblock \href {https://arxiv.org/abs/2401.14196} {Deepseek-coder: When the large language model meets programming -- the rise of code intelligence}.
\newblock \emph{Preprint}, arXiv:2401.14196.

\bibitem[{Guo et~al.(2024{\natexlab{b}})Guo, Liu, Zhang, and Wang}]{guo2024can}
Hongyi Guo, Zhihan Liu, Yufeng Zhang, and Zhaoran Wang. 2024{\natexlab{b}}.
\newblock Can large language models play games? a case study of a self-play approach.
\newblock \emph{arXiv preprint arXiv:2403.05632}.

\bibitem[{Hendrycks et~al.(2021)Hendrycks, Burns, Kadavath, Arora, Basart, Tang, Song, and Steinhardt}]{MATH}
Dan Hendrycks, Collin Burns, Saurav Kadavath, Akul Arora, Steven Basart, Eric Tang, Dawn Song, and Jacob Steinhardt. 2021.
\newblock Measuring mathematical problem solving with the math dataset.
\newblock \emph{NeurIPS}.

\bibitem[{Huang et~al.(2024)Huang, Liu, Chen, Wang, Wang, Lian, Wang, Tang, and Chen}]{UnderstandingThePlanning_Huang2024}
Xu~Huang, Weiwen Liu, Xiaolong Chen, Xingmei Wang, Hao Wang, Defu Lian, Yasheng Wang, Ruiming Tang, and Enhong Chen. 2024.
\newblock Understanding the planning of llm agents: A survey.
\newblock \emph{arXiv preprint arXiv:2402.02716}.

\bibitem[{Kwon et~al.(2023)Kwon, Li, Zhuang, Sheng, Zheng, Yu, Gonzalez, Zhang, and Stoica}]{EfficientMemoryManagement_Kwon2023}
Woosuk Kwon, Zhuohan Li, Siyuan Zhuang, Ying Sheng, Lianmin Zheng, Cody~Hao Yu, Joseph~E. Gonzalez, Hao Zhang, and Ion Stoica. 2023.
\newblock Efficient memory management for large language model serving with pagedattention.
\newblock In \emph{Proceedings of the ACM SIGOPS 29th Symposium on Operating Systems Principles}.

\bibitem[{Leviathan et~al.(2023)Leviathan, Kalman, and Matias}]{leviathan2023fastinferencetransformersspeculative}
Yaniv Leviathan, Matan Kalman, and Yossi Matias. 2023.
\newblock \href {https://arxiv.org/abs/2211.17192} {Fast inference from transformers via speculative decoding}.
\newblock \emph{Preprint}, arXiv:2211.17192.

\bibitem[{Li et~al.(2024{\natexlab{a}})Li, Wei, Zhang, and Zhang}]{eagle2}
Yuhui Li, Fangyun Wei, Chao Zhang, and Hongyang Zhang. 2024{\natexlab{a}}.
\newblock \href {https://arxiv.org/abs/2406.16858} {Eagle-2: Faster inference of language models with dynamic draft trees}.
\newblock \emph{Preprint}, arXiv:2406.16858.

\bibitem[{Li et~al.(2024{\natexlab{b}})Li, Wei, Zhang, and Zhang}]{eagle}
Yuhui Li, Fangyun Wei, Chao Zhang, and Hongyang Zhang. 2024{\natexlab{b}}.
\newblock \href {https://arxiv.org/abs/2401.15077} {Eagle: Speculative sampling requires rethinking feature uncertainty}.
\newblock \emph{Preprint}, arXiv:2401.15077.

\bibitem[{Li et~al.(2025)Li, Wei, Zhang, and Zhang}]{li2025eagle3scalinginferenceacceleration}
Yuhui Li, Fangyun Wei, Chao Zhang, and Hongyang Zhang. 2025.
\newblock \href {https://arxiv.org/abs/2503.01840} {{EAGLE-3}: Scaling up inference acceleration of large language models via training-time test}.
\newblock \emph{Preprint}, arXiv:2503.01840.

\bibitem[{Lin et~al.(2024{\natexlab{a}})Lin, Tang, Tang, Yang, Chen, Wang, Xiao, Dang, Gan, and Han}]{awq}
Ji~Lin, Jiaming Tang, Haotian Tang, Shang Yang, Wei-Ming Chen, Wei-Chen Wang, Guangxuan Xiao, Xingyu Dang, Chuang Gan, and Song Han. 2024{\natexlab{a}}.
\newblock \href {https://arxiv.org/abs/2306.00978} {Awq: Activation-aware weight quantization for llm compression and acceleration}.
\newblock \emph{Preprint}, arXiv:2306.00978.

\bibitem[{Lin et~al.(2024{\natexlab{b}})Lin, Tang, Yang, Zhang, Xiao, Gan, and Han}]{qserve}
Yujun Lin, Haotian Tang, Shang Yang, Zhekai Zhang, Guangxuan Xiao, Chuang Gan, and Song Han. 2024{\natexlab{b}}.
\newblock \href {https://arxiv.org/abs/2405.04532} {Qserve: W4a8kv4 quantization and system co-design for efficient llm serving}.
\newblock \emph{Preprint}, arXiv:2405.04532.

\bibitem[{Liu et~al.(2024)Liu, Hu, Bailis, Cheung, Deng, Stoica, and Zhang}]{liu2024onlinespeculativedecoding}
Xiaoxuan Liu, Lanxiang Hu, Peter Bailis, Alvin Cheung, Zhijie Deng, Ion Stoica, and Hao Zhang. 2024.
\newblock \href {https://arxiv.org/abs/2310.07177} {Online speculative decoding}.
\newblock \emph{Preprint}, arXiv:2310.07177.

\bibitem[{Merity et~al.(2016)Merity, Xiong, Bradbury, and Socher}]{wikitext2}
Stephen Merity, Caiming Xiong, James Bradbury, and Richard Socher. 2016.
\newblock \href {https://arxiv.org/abs/1609.07843} {Pointer sentinel mixture models}.
\newblock \emph{Preprint}, arXiv:1609.07843.

\bibitem[{Miao et~al.(2024)Miao, Oliaro, Zhang, Cheng, Wang, Zhang, Wong, Zhu, Yang, Shi, Shi, Chen, Arfeen, Abhyankar, and Jia}]{spec_infer}
Xupeng Miao, Gabriele Oliaro, Zhihao Zhang, Xinhao Cheng, Zeyu Wang, Zhengxin Zhang, Rae Ying~Yee Wong, Alan Zhu, Lijie Yang, Xiaoxiang Shi, Chunan Shi, Zhuoming Chen, Daiyaan Arfeen, Reyna Abhyankar, and Zhihao Jia. 2024.
\newblock \href {https://doi.org/10.1145/3620666.3651335} {Specinfer: Accelerating large language model serving with tree-based speculative inference and verification}.
\newblock In \emph{Proceedings of the 29th ACM International Conference on Architectural Support for Programming Languages and Operating Systems, Volume 3}, ASPLOS ’24. ACM.

\bibitem[{{PyTorch Contributors}(2024)}]{pytorch2024reproducibility}
{PyTorch Contributors}. 2024.
\newblock Reproducibility.
\newblock \url{https://pytorch.org/docs/stable/notes/randomness.html}.
\newblock PyTorch Documentation, Accessed: January 2024.

\bibitem[{Rein et~al.(2023)Rein, Hou, Stickland, Petty, Pang, Dirani, Michael, and Bowman}]{rein2023gpqagraduatelevelgoogleproofqa}
David Rein, Betty~Li Hou, Asa~Cooper Stickland, Jackson Petty, Richard~Yuanzhe Pang, Julien Dirani, Julian Michael, and Samuel~R. Bowman. 2023.
\newblock \href {https://arxiv.org/abs/2311.12022} {Gpqa: A graduate-level google-proof qa benchmark}.
\newblock \emph{Preprint}, arXiv:2311.12022.

\bibitem[{RyokoAI(2021)}]{sharegpt52k}
RyokoAI. 2021.
\newblock Sharegpt52k.

\bibitem[{Sakaguchi et~al.(2019)Sakaguchi, Bras, Bhagavatula, and Choi}]{ai2:winogrande}
Keisuke Sakaguchi, Ronan~Le Bras, Chandra Bhagavatula, and Yejin Choi. 2019.
\newblock \href {https://arxiv.org/abs/1907.10641} {Winogrande: An adversarial winograd schema challenge at scale}.
\newblock \emph{Preprint}, arXiv:1907.10641.

\bibitem[{Shao et~al.(2024)Shao, Wang, Zhu, Xu, Song, Bi, Zhang, Zhang, Li, Wu, and Guo}]{ds_math}
Zhihong Shao, Peiyi Wang, Qihao Zhu, Runxin Xu, Junxiao Song, Xiao Bi, Haowei Zhang, Mingchuan Zhang, Y.~K. Li, Y.~Wu, and Daya Guo. 2024.
\newblock \href {https://arxiv.org/abs/2402.03300} {Deepseekmath: Pushing the limits of mathematical reasoning in open language models}.
\newblock \emph{Preprint}, arXiv:2402.03300.

\bibitem[{Touvron et~al.(2023)Touvron, Martin, Stone, Albert, Almahairi, Babaei, Bashlykov, Batra, Bhargava, Bhosale, Bikel, Blecher, Ferrer, Chen, Cucurull, Esiobu, Fernandes, Fu, Fu, Fuller, Gao, Goswami, Goyal, Hartshorn, Hosseini, Hou, Inan, Kardas, Kerkez, Khabsa, Kloumann, Korenev, Koura, Lachaux, Lavril, Lee, Liskovich, Lu, Mao, Martinet, Mihaylov, Mishra, Molybog, Nie, Poulton, Reizenstein, Rungta, Saladi, Schelten, Silva, Smith, Subramanian, Tan, Tang, Taylor, Williams, Kuan, Xu, Yan, Zarov, Zhang, Fan, Kambadur, Narang, Rodriguez, Stojnic, Edunov, and Scialom}]{llama2}
Hugo Touvron, Louis Martin, Kevin Stone, Peter Albert, Amjad Almahairi, Yasmine Babaei, Nikolay Bashlykov, Soumya Batra, Prajjwal Bhargava, Shruti Bhosale, Dan Bikel, Lukas Blecher, Cristian~Canton Ferrer, Moya Chen, Guillem Cucurull, David Esiobu, Jude Fernandes, Jeremy Fu, Wenyin Fu, and 49 others. 2023.
\newblock \href {https://arxiv.org/abs/2307.09288} {Llama 2: Open foundation and fine-tuned chat models}.
\newblock \emph{Preprint}, arXiv:2307.09288.

\bibitem[{Wang et~al.(2023)Wang, Ivison, Dasigi, Hessel, Khot, Chandu, Wadden, MacMillan, Smith, Beltagy, and Hajishirzi}]{wang2023far}
Yizhong Wang, Hamish Ivison, Pradeep Dasigi, Jack Hessel, Tushar Khot, Khyathi~Raghavi Chandu, David Wadden, Kelsey MacMillan, Noah~A. Smith, Iz~Beltagy, and Hannaneh Hajishirzi. 2023.
\newblock \href {https://arxiv.org/abs/2306.04751} {How far can camels go? exploring the state of instruction tuning on open resources}.
\newblock \emph{Preprint}, arXiv:2306.04751.

\bibitem[{Xiao et~al.(2023)Xiao, Lin, Seznec, Wu, Demouth, and Han}]{xiao2023smoothquant}
Guangxuan Xiao, Ji~Lin, Mickael Seznec, Hao Wu, Julien Demouth, and Song Han. 2023.
\newblock Smoothquant: Accurate and efficient post-training quantization for large language models.
\newblock In \emph{International Conference on Machine Learning}.

\bibitem[{Xiong et~al.(2024)Xiong, Shi, Shen, Rosenberg, Qin, Calandriello, Khalman, Joshi, Piot, Saleh, Jin, Zhang, and Liu}]{xiong2024buildingmathagentsmultiturn}
Wei Xiong, Chengshuai Shi, Jiaming Shen, Aviv Rosenberg, Zhen Qin, Daniele Calandriello, Misha Khalman, Rishabh Joshi, Bilal Piot, Mohammad Saleh, Chi Jin, Tong Zhang, and Tianqi Liu. 2024.
\newblock \href {https://arxiv.org/abs/2409.02392} {Building math agents with multi-turn iterative preference learning}.
\newblock \emph{Preprint}, arXiv:2409.02392.

\bibitem[{Yan et~al.(2025)Yan, Agarwal, and Venkataraman}]{yan2025decodingspeculativedecoding}
Minghao Yan, Saurabh Agarwal, and Shivaram Venkataraman. 2025.
\newblock \href {https://arxiv.org/abs/2402.01528} {Decoding speculative decoding}.
\newblock \emph{Preprint}, arXiv:2402.01528.

\bibitem[{Ye et~al.(2025)Ye, Chen, Lai, Lin, Zhang, Wang, Chen, Kasikci, Grover, Krishnamurthy, and Ceze}]{flashinfer}
Zihao Ye, Lequn Chen, Ruihang Lai, Wuwei Lin, Yineng Zhang, Stephanie Wang, Tianqi Chen, Baris Kasikci, Vinod Grover, Arvind Krishnamurthy, and Luis Ceze. 2025.
\newblock \href {https://arxiv.org/abs/2501.01005} {Flashinfer: Efficient and customizable attention engine for llm inference serving}.
\newblock \emph{Preprint}, arXiv:2501.01005.

\bibitem[{Yu et~al.(2022)Yu, Jeong, Kim, Kim, and Chun}]{orca}
Gyeong-In Yu, Joo~Seong Jeong, Geon-Woo Kim, Soojeong Kim, and Byung-Gon Chun. 2022.
\newblock \href {https://www.usenix.org/conference/osdi22/presentation/yu} {Orca: A distributed serving system for {Transformer-Based} generative models}.
\newblock In \emph{16th USENIX Symposium on Operating Systems Design and Implementation (OSDI 22)}, pages 521--538, Carlsbad, CA. USENIX Association.

\bibitem[{Zhang et~al.(2023)Zhang, Press, Merrill, Liu, and Smith}]{HalluSnowball}
Muru Zhang, Ofir Press, William Merrill, Alisa Liu, and Noah~A Smith. 2023.
\newblock How language model hallucinations can snowball.
\newblock \emph{arXiv preprint arXiv:2305.13534}.

\bibitem[{Zhao et~al.(2024{\natexlab{a}})Zhao, Wan, Peng, Lin, and Wu}]{zhao2024llmpqservingllmheterogeneous}
Juntao Zhao, Borui Wan, Yanghua Peng, Haibin Lin, and Chuan Wu. 2024{\natexlab{a}}.
\newblock \href {https://arxiv.org/abs/2403.01136} {Llm-pq: Serving llm on heterogeneous clusters with phase-aware partition and adaptive quantization}.
\newblock \emph{Preprint}, arXiv:2403.01136.

\bibitem[{Zhao et~al.(2024{\natexlab{b}})Zhao, Lin, Zhu, Ye, Chen, Zheng, Ceze, Krishnamurthy, Chen, and Kasikci}]{Atom}
Yilong Zhao, Chien-Yu Lin, Kan Zhu, Zihao Ye, Lequn Chen, Size Zheng, Luis Ceze, Arvind Krishnamurthy, Tianqi Chen, and Baris Kasikci. 2024{\natexlab{b}}.
\newblock \href {https://arxiv.org/abs/2310.19102} {Atom: Low-bit quantization for efficient and accurate llm serving}.
\newblock \emph{Preprint}, arXiv:2310.19102.

\bibitem[{Zheng et~al.(2023{\natexlab{a}})Zheng, Chiang, Sheng, Li, Zhuang, Wu, Zhuang, Li, Lin, Xing et~al.}]{lmsys}
Lianmin Zheng, Wei-Lin Chiang, Ying Sheng, Tianle Li, Siyuan Zhuang, Zhanghao Wu, Yonghao Zhuang, Zhuohan Li, Zi~Lin, Eric Xing, and 1 others. 2023{\natexlab{a}}.
\newblock Lmsys-chat-1m: A large-scale real-world llm conversation dataset.
\newblock \emph{arXiv preprint arXiv:2309.11998}.

\bibitem[{Zheng et~al.(2023{\natexlab{b}})Zheng, Chiang, Sheng, Zhuang, Wu, Zhuang, Lin, Li, Li, Xing, Zhang, Gonzalez, and Stoica}]{zheng2023judgingllmasajudgemtbenchchatbot}
Lianmin Zheng, Wei-Lin Chiang, Ying Sheng, Siyuan Zhuang, Zhanghao Wu, Yonghao Zhuang, Zi~Lin, Zhuohan Li, Dacheng Li, Eric~P. Xing, Hao Zhang, Joseph~E. Gonzalez, and Ion Stoica. 2023{\natexlab{b}}.
\newblock \href {https://arxiv.org/abs/2306.05685} {Judging llm-as-a-judge with mt-bench and chatbot arena}.
\newblock \emph{Preprint}, arXiv:2306.05685.

\end{thebibliography}

\newpage
\appendix
\clearpage
\section{Full Experiment Result.}~\label{sec:app_full_res}
\subsection{Acceleration Evaluation}
\begin{table*}[!t]

\centering
\small
\caption{
Comparison of token generation throughput across different model sizes, quantization configurations, and batch sizes for various datasets. All values are measured in token/s.
“Avg.” denotes the average speedup ratio for the corresponding row or column. 
}
\resizebox{\textwidth}{!}{%
\begin{tabular}{cccccccccc}
\toprule
\textbf{Model} & \textbf{Method} & \textbf{Batch} & \textbf{GSM8K} & \textbf{MATH} & \textbf{MBPP} & \textbf{HumanEval} & \textbf{ShareGPT} & \textbf{LMsys-1k} & \textbf{Avg.} \\
\midrule
\multirow{13}{*}{3B\footnotemark[1]}
&  \multirow{3}{*}{W16A16} 
& 8 & 511.1 & 588.7 & 756.6 & 647.2 & 785.7 & 711.2 & -- \\
& & 16 & 666.5 & 845.6 & 1171.0 & 948.3 & 1292.2 & 1126.4 & -- \\
& & 32 & 833.4 & 1081.5 & 1697.7 & 1111.6 & 1975.6 & 1553.3 & --  \\
\cmidrule{2-10}
&  \multirow{3}{*}{W4A4} 
& 8 & 804.7 & 921.2 & 1002.0 & 892.6 & 1091.6 & 990.3 & -- \\
& & 16 & 1109.1 & 1374.5 & 1548.0 & 1289.8 & 1763.5 & 1581.0 & -- \\
& & 32 & 1424.3 & 1899.3 & 2300.6 & 1488.2 & 2777.3 & 2194.4 & --  \\
\cmidrule{2-10}
& \multirow{3}{*}{W4A16} 
& 8 & 420.0 & 476.7 & 604.5 & 535.7 & 610.4 & 559.8 & -- \\
& & 16 & 578.5 & 715.9 & 989.7 & 804.4 & 1080.2 & 925.8 & -- \\
& & 32 & 726.3 & 933.8 & 1536.7 & 954.4 & 1704.5 & 1336.4 & -- \\
\cmidrule{2-10}
&  \multirow{4}{*}{\sys{}} 
& 8 & 594.1 (1.41$\times$) & 648.2 (1.36$\times$) & 760.1 (1.26$\times$) & 723.6 (1.35$\times$) & 787.5 (1.29$\times$) & 738.8 (1.32$\times$) & 1.33$\times$ \\
& & 16 & 811.5 (1.40$\times$) & 936.0 (1.31$\times$) & 1157.8 (1.17$\times$) & 1042.1 (1.30$\times$) & 1294.5 (1.20$\times$) & 1171.4 (1.27$\times$) & 1.27$\times$ \\
& & 32 & 1030.4 (1.42$\times$) & 1240.2 (1.33$\times$) & 1617.4 (1.05$\times$) & 1248.5 (1.31$\times$) & 1969.6 (1.16$\times$) & 1576.0 (1.18$\times$) & 1.24$\times$  \\
\cmidrule{3-10}
& & Avg. & 1.41$\times$ & 1.33$\times$ & 1.16$\times$ & 1.32$\times$ & 1.21$\times$ & 1.25 $\times$ & 1.28$\times$ \\
\midrule
\multirow{13}{*}{7B} &  \multirow{3}{*}{W16A16} 
& 8 & 213.4 & 254.3 & 278.8 & 316.7 & 322.4 & 285.3 & -- \\
& & 16 & 290.3 & 362.1 & 447.7 & 505.1 & 541.3 & 441.6 & -- \\
& & 32 & 340.9 & 441.6 & 585.3 & 663.6 & 735.3 & 564.2 & -- \\
\cmidrule{2-10}
&  \multirow{3}{*}{W4A4} 
& 8 & 349.5 & 411.7 & 396.1 & 471.2 & 471.8 & 419.4 & -- \\
& & 16 & 496.6 & 612.2 & 614.3 & 749.5 & 760.9 & 642.6 & -- \\
& & 32 & 620.0 & 793.6 & 801.5 & 1043.9 & 1083.2 & 865.5 & --  \\
\cmidrule{2-10}
& \multirow{3}{*}{W4A16} 
& 8 & 165.0 & 193.1 & 224.5 & 240.2 & 243.5 & 220.2 & -- \\
& & 16 & 231.8 & 286.5 & 384.4 & 407.3 & 435.9 & 358.0 & -- \\
& & 32 & 268.9 & 359.9 & 480.0 & 555.9 & 620.2 & 470.1 & -- \\
\cmidrule{2-10}
&  \multirow{4}{*}{\sys{}} 
& 8 & 253.7 (1.54$\times$) & 291.5 (1.51$\times$) & 298.3 (1.33$\times$) & 350.9 (1.46$\times$) & 345.7 (1.42$\times$) & 310.3 (1.41$\times$) & 1.44$\times$ \\
& & 16 & 359.8 (1.55$\times$) & 420.2 (1.47$\times$) & 466.7 (1.21$\times$) & 555.2 (1.36$\times$) & 557.8 (1.28$\times$) & 473.1 (1.32$\times$) & 1.37$\times$ \\
& & 32 & 441.8 (1.64$\times$) & 527.2 (1.46$\times$) & 575.3 (1.20$\times$) & 749.4 (1.35$\times$) & 770.0 (1.24$\times$) & 628.4 (1.34$\times$) & 1.39$\times$ \\
\cmidrule{3-10}
& & Avg. & 1.58$\times$ & 1.48$\times$ & 1.25$\times$ & 1.39$\times$ & 1.31$\times$ & 1.36$\times$ & 1.39$\times$ \\
\midrule
\multirow{13}{*}{8B}&  \multirow{3}{*}{W16A16} 
& 8 & 189.4 & 211.5 & 256.0 & 259.1 & 290.7 & 265.8 & -- \\
& & 16 & 262.0 & 311.2 & 408.7 & 401.2 & 511.0 & 447.4 & -- \\
& & 32 & 303.8 & 390.8 & 566.3 & 522.6 & 820.0 & 649.8 & --  \\
\cmidrule{2-10}
&  \multirow{3}{*}{W4A4} 
& 8 & 295.3 & 323.5 & 344.6 & 354.4 & 395.9 & 366.8 & -- \\
& & 16 & 431.4 & 503.3 & 536.8 & 566.4 & 697.5 & 621.1 & -- \\
& & 32 & 532.8 & 688.5 & 755.7 & 763.7 & 1167.9 & 956.8 & --  \\
\cmidrule{2-10}
& \multirow{3}{*}{W4A16} 
& 8 & 155.6 & 173.8 & 215.0 & 208.7 & 231.1 & 215.6 & -- \\
& & 16 & 222.9 & 263.0 & 354.8 & 345.9 & 422.8 & 369.4 & -- \\
& & 32 & 299.3 & 363.3 &  509.8 & 468.7 & 706.0 & 580.5 & -- \\
\cmidrule{2-10}
&  \multirow{4}{*}{\sys{}} 
& 8 & 222.6 (1.43$\times$) & 233.9 (1.35$\times$) & 256.7 (1.19$\times$) & 271.5 (1.30$\times$) & 285.0 (1.23$\times$) & 268.3 (1.24$\times$) & 1.29$\times$ \\
& & 16 & 322.6 (1.45$\times$) & 362.5 (1.38$\times$) & 402.7 (1.14$\times$) & 438.5 (1.27$\times$) & 507.5 (1.20$\times$) & 453.5 (1.23$\times$) & 1.28$\times$ \\
& & 32 & 400.2 (1.34$\times$) & 483.0 (1.33$\times$) & 578.1 (1.13$\times$) & 573.0 (1.22$\times$) & 798.8 (1.13$\times$) & 684.5 (1.18$\times$) & 1.27$\times$ \\
\cmidrule{3-10}
& & Avg. &1.44$\times$ & 1.36$\times$ & 1.15$\times$ & 1.26$\times$ & 1.19$\times$ & 1.22$\times$ & 1.27 $\times$ \\
\midrule
\multirow{13}{*}{13B\footnotemark[1]} 
&  \multirow{3}{*}{W16A16} 
& 8 & 121.9 & 146.6 & 183.1 & 182.0 & 187.1 & 160.1 & -- \\
& & 16 & 169.6 & 211.2 & 304.4 & 291.0 & 311.0 & 243.0 & -- \\
& & 32 & 202.4 & 253.8 & 426.0 & 423.5 & 311.0 & 334.2 & --  \\
\cmidrule{2-10}
&  \multirow{3}{*}{W4A4} 
& 8 & 194.7 & 228.2 & 253.6 & 261.5 & 259.8 & 228.2 & -- \\
& & 16 & 288.3 & 349.2 & 415.3 & 424.9 & 431.5 & 348.4 & -- \\
& & 32 & 369.8 & 469.9 & 606.7 & 665.4 & 431.5 & 508.8 & --  \\
\cmidrule{2-10}
& \multirow{3}{*}{W4A16} 
& 8 & 94.8 & 112.9 & 143.4 & 140.0 & 146.7 & 127.9 & -- \\
& & 16 & 136.1 & 171.9 & 250.8 & 236.9 & 255.9 & 207.2 & -- \\
& & 32 &  207.5 & 241.6 & 376.4 & 365.5 & 255.9 & 287.4 & -- \\
\cmidrule{2-10}
&  \multirow{4}{*}{\sys{}} 
& 8 & 148.2 (1.56$\times$) & 167.9 (1.49$\times$) & 193.6 (1.35$\times$) & 201.2 (1.44$\times$) & 194.5 (1.33$\times$) & 174.0 (1.36$\times$) & 1.42$\times$ \\
& & 16 & 212.8 (1.56$\times$) & 248.6 (1.45$\times$) & 316.8 (1.26$\times$) & 323.3 (1.36$\times$) & 327.4 (1.28$\times$) & 266.9 (1.29$\times$) & 1.29$\times$ \\
& & 32 & 266.6 (1.28$\times$) & 320.0 (1.32$\times$) & 451.5 (1.20$\times$) & 483.0 (1.32$\times$) & 327.4 (1.28$\times$) & 379.3 (1.32$\times$)  & 1.32$\times$  \\
\cmidrule{3-10}
& & Avg. & 1.56$\times$ & 1.47$\times$ & 1.27$\times$ & 1.37$\times$ & 1.29$\times$ & 1.32$\times$ & 1.38$\times$ \\
\bottomrule
\end{tabular}%
}
\label{tab:full_performance_comparison_l20}
\end{table*}
\textbf{Versus quantization configurations.}
Table~\ref{tab:full_performance_comparison_l20} presents the comprehensive comparison of token generation throughput across multiple dimensions: model sizes, quantization configurations, and batch sizes, evaluated on various datasets. It is noteworthy that Llama2-7B shows higher speedup than Llama3-8B. This stems from the size difference primarily related to vocabulary, coupled with the introduction of Group-Query Attention~\citep{GQA}, reducing the computation workload.


\begin{table*}[!t]
\small
\centering
\caption{Performance comparison of EAGLE-Quant, \sys, W4A16, and W4A4 on Llama-2-7b-chat-hf across different batch sizes. Results are reported for GSM8K (8-shot), MATH (4-shot), MBPP (0-shot), HumanEval (0-shot), ShareGPT, and LMsys-1k benchmarks. ``OOM'' indicates out-of-memory errors. Better case for \sys{} or EAGLE is marked in gray. In the case of batch size=8, the speedup ratio of \sys{} compared to EAGLE is indicated in parentheses next to the data points.}
\label{tab:full_spec}
\resizebox{\textwidth}{!}{%
\begin{tabular}{lccccccc}
\toprule
\makecell{\textbf{Method}}
& \textbf{Batch Size}
& \textbf{GSM8K (8-shot)}
& \textbf{MATH (4-shot)}
& \textbf{MBPP (0-shot)}
& \textbf{HumanEval (0-shot)}
& \textbf{ShareGPT}
& \textbf{LMsys-1k} \\
\midrule

\rowcolor{gray!15}
\multirow{3}{*}{EAGLE} 
& 1  & 65.81  & 70.11  & 68.53  & 49.15  & 79.60  & 71.29 \\
& 8  & 140.16 & 210.52 & 138.01 & 136.86 & 247.42 & 167.57 \\
& 16 & OOM    & OOM    & OOM    & OOM    & OOM    & OOM \\
\midrule

& 1  & 51.25  & 48.36  & 56.87  & 54.22  & 56.77  & 56.14 \\
\rowcolor{gray!15}
\sys{} & 8  & 208.95 (1.49$\times$) & 249.97 (1.19$\times$) & 185.13 (1.34$\times$) & 185.99 (1.36$\times$) & 329.44 (1.33$\times$) & 260.48 (1.55$\times$) \\
\rowcolor{gray!15}
& 16 & 292.82 & 356.93 & 269.87 & 255.11 & 562.07 & 463.35 \\
\midrule

\multirow{3}{*}{W4A16} 
& 1  & 59.80  & 63.65  & 75.49  & 72.04  & 76.04  & 72.27 \\
& 8  & 146.34 & 185.52 & 180.12 & 163.54 & 250.06 & 213.66 \\
& 16 & 190.09 & 251.59 & 254.24 & 211.49 & 458.57 & 371.38 \\
\midrule

\multirow{3}{*}{W4A4} 
& 1  & 64.79  & 66.32  & 74.72  & 73.47  & 73.09  & 71.55 \\
& 8  & 284.84 & 369.27 & 250.11 & 256.54 & 492.21 & 393.12 \\
& 16 & 401.77 & 540.82 & 357.62 & 330.71 & 895.77 & 713.67 \\
\bottomrule
\end{tabular}%
}
\end{table*}

\textbf{Versus speculative decoding.} 
Table~\ref{tab:full_spec} presents a comprehensive comparison between our approach and the EAGLE method across multiple benchmarks.

\subsection{Ablation Studies}~\label{sec:app_comphre}

\begin{figure*}[!t]
    \centering
    \makebox[\textwidth]{ 
        \begin{minipage}{0.48\textwidth}
            \centering
            \includegraphics[width=\linewidth]{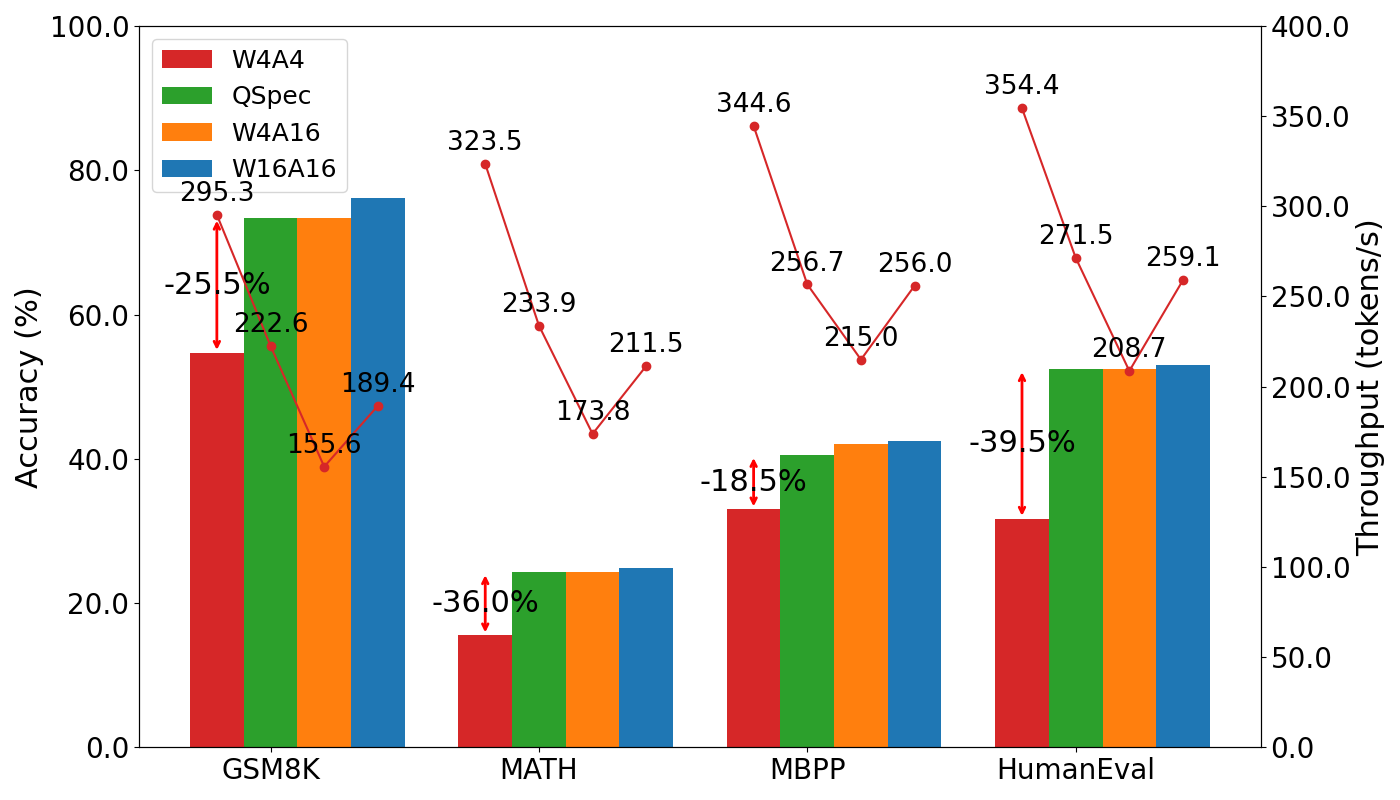}
            {\small Batch Size 8} 
        \end{minipage}
        \hspace{0.02\textwidth}
        \begin{minipage}{0.48\textwidth}
            \centering
            \includegraphics[width=\linewidth]{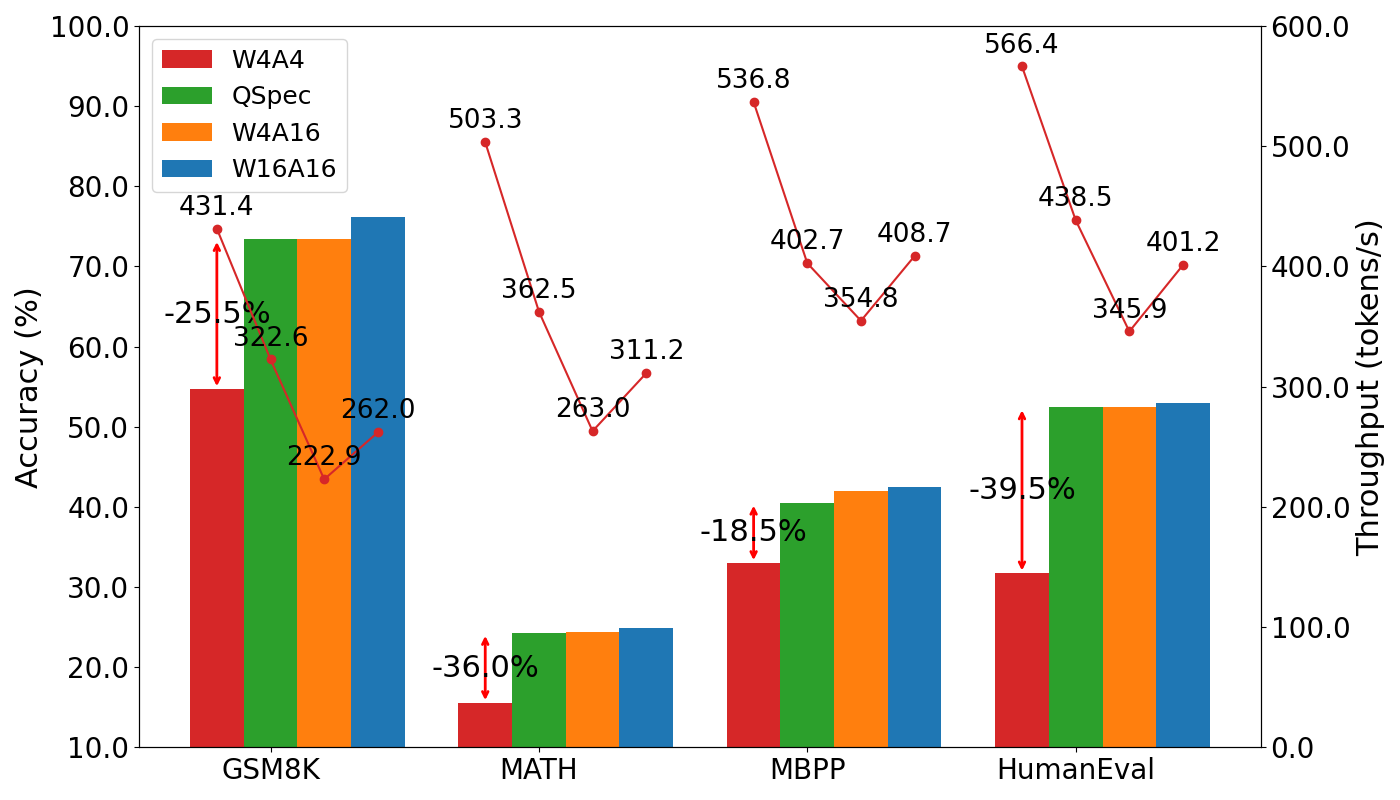}
            {\small Batch Size 16} 
        \end{minipage}
    }
    \caption{Comparison of accuracy and efficiency among W16A16, W4A16, W4A4, and \sys{} across various datasets with batch sizes of 8 and 16 on Llama3-8b-instruct model. The bars and lines represent the accuracy and throughput of each method.}
    \label{fig:acc_throughput}
\end{figure*}

\noindent
\textbf{Performance Trade-offs: \sys{} versus W4A16.}
We conduct a thorough analysis of the trade-offs between throughput and accuracy for our proposed framework against all baseline implementations. Figure~\ref{fig:acc_throughput} illustrates this comparison, plotting generation quality (accuracy) against computational efficiency (throughput). Our analysis reveals that while W4A4 suffers substantial performance degradation (18.5\%-39.5\% reduction) on multi-step reasoning benchmarks compared to W4A16, \sys{} achieves comparable accuracy to W4A16 while delivering significantly higher throughput. Although \sys{}'s accuracy is marginally lower than W16A16 due to weight quantization-induced memory optimization, it successfully preserves the performance characteristics of W4A16 while offering superior computational efficiency.

\subsection{Datasets and Sampling}\label{app:datasets} For the PIQA and Winogrande datasets, we randomly select 500 questions from each for performance evaluation. In contrast, we process the entire GSM8K and MATH datasets, as detailed in Atom~\citet{Atom}. When adapting \sys{} to the Quarot method, we sample 200 samples from the GSM8K dataset, while sampling 700 from the MATH dataset, ensuring balanced representation by selecting 100 questions from each of the seven distinct question types.
Additionally, we sample 200 questions from the MBPP dataset, while processing the entire HumanEval dataset. This preserves a thorough assessment of the model's performance across various datasets. For acceleration evaluation, we maintain the random seed at 42, and sample 100 samples from the GSM8K, MATH, MBPP, HumanEval, ShareGPT, and LMSys datasets, respsectively. This consistent methodology guarantees that our evaluation remains reproducible and representative across these diverse datasets.

\subsection{\sys{} on vLLM: Real-World Serving Evaluation} \label{app:vLLM}

To validate \sys{}'s effectiveness in real-world serving scenarios, we integrated it into vLLM\footnote{We implemented \sys{} on Commit 9a7c3a0 of vLLM from 21 Jan. 2025}\cite{EfficientMemoryManagement_Kwon2023} by developing custom kernels and a tailored vLLM worker, enabling shared weights and KV rewriting. We conducted performance tests on Llama-3-8b-instruct model with five diverse test sets: Wild Chat\cite{deng2024wildvisopensourcevisualizer}, GSM8K\cite{gsm8k}, MBPP\cite{MBPP}, MT-Bench\cite{zheng2023judgingllmasajudgemtbenchchatbot}, and GPQA-Diamond\cite{rein2023gpqagraduatelevelgoogleproofqa}---covering domains such as chat, mathematics, coding, and general knowledge. Note that vLLM's support for the speculative decoding paradigm remains suboptimal\footnote{For more detailed explanation, please refer to https://docs.vllm.ai/en/stable/features/spec\_decode.html.} due to its complex scheduling and memory management mechanisms. Our experiments were performed on an NVIDIA A100 GPU (40GB), with batch sizes ranging from 1 to 32, a draft length $\gamma=3$, and a baseline of autoregressive decoding using W4A16 with same weights. Results are shown in Table~\ref{tab:vllm}. \sys{} achieves an average speedup of 1.24$\times$, maintaining effective acceleration even at a batch size of 32---a challenging feat for prior work. Additionally, we report acceptance rates across these test sets, with \sys{} achieving an impressive 93\%--95\% acceptance rate. This aligns with our findings in the Section~\ref{sec: motivation}, suggesting that leveraging these high acceptance rates for further acceleration is a promising direction for future \sys{} research.

Our integration of \sys{} into vLLM also initially aimed to enable a fair comparison with EAGLE~\citep{eagle}. Unfortunately, we could not reproduce EAGLE's performance on vLLM, as it exhibited significant performance degradation in batched scenarios, even losing speedup in the W4A16 setting. Moreover, due to the poor performance and inconsistent compatibility of other speculative decoding methods on vLLM, our additional experiments here do not include comparisons with other methods. We also note that the EAGLE team released EAGLE-3\citep{li2025eagle3scalinginferenceacceleration} in late March 2025, open-sourcing partial pretrained model weights in April and claiming competitive speedup on vLLM. However, as of now, the code to reproduce this claim on vLLM is unavailable, preventing us from evaluating EAGLE-3's performance on vLLM. Thus, we exclude comparisons with EAGLE-3, in line with the guidelines on handling contemporaneous work. 
\noindent
This does not detract from the fact that both \sys{} and EAGLE-3 represent outstanding contributions developed concurrently. As discussed in Section~\ref{app:limit}, \sys{} and EAGLE exemplify two distinct design paradigms: EAGLE follows a training-intensive route with an independently optimized draft model and a refined acceptance policy, while \sys{} adopts a training-free framework that leverages shared weights and KV cache in a unified structure. Each approach has its own strengths, and we look forward to future opportunities for dialogue and collaboration with the EAGLE team.


\begin{table}[h]
\centering
\caption{Performance of \sys{} on vLLM across different batch sizes and test sets, with acceptance rates per test set respectively. All speedup values are reported with a $\times$ multiplier.}
\label{tab:vllm}
\small
\renewcommand{\arraystretch}{1.3} 
\resizebox{1\columnwidth}{!}{%
\begin{tabular}{m{2.5cm} *{6}{S[table-format=1.2, table-space-text-post=$\times$]} S[table-format=2.1, table-column-width=2.5cm]}
\toprule
\multirow{2}{2.5cm}{\centering \textbf{Test Set}} & \multicolumn{6}{c}{\multirow{1.5}{*}{\centering \textbf{Batch Size}}} & \multicolumn{1}{c}{\multirow{2}{2.5cm}{\centering \textbf{Acceptance Rate (\%)}}} \\
\cmidrule(lr){2-7}
 & {1} & {2} & {4} & {8} & {16} & {32} & \\
\midrule
Wild Chat & 1.29 $\times$ & 1.33 $\times$ & 1.29 $\times$ & 1.26 $\times$ & 1.28 $\times$ & 1.13 $\times$ & 93.1 \\
GSM8K & 1.36 $\times$ & 1.33 $\times$ & 1.23 $\times$ & 1.24 $\times$ & 1.25 $\times$ & 1.01 $\times$ & 92.1 \\
MBPP & 1.34 $\times$ & 1.26 $\times$ & 1.16 $\times$ & 1.18 $\times$ & 1.18 $\times$ & 1.12 $\times$ & 95.5 \\
MT-Bench & 1.33 $\times$ & 1.24 $\times$ & 1.28 $\times$ & 1.30 $\times$ & 1.16 $\times$ & 1.15 $\times$ & 93.5 \\
GPQA-Diamond & 1.33 $\times$ & 1.31 $\times$ & 1.21 $\times$ & 1.18 $\times$ & 1.23 $\times$ & 1.10 $\times$ & 94.1 \\
\bottomrule
\end{tabular}
}
\end{table}

\subsection{Artifact Documentation}

We provide the official implementation of \sys{} in the supplementary materials. The codebase is fully documented and includes\footnote{https://github.com/hku-netexplo-lab/QSpec}:

\begin{itemize}
    \item \textbf{Installation guide} covering dependency setup (CUDA 12.5, Python 3.10), environment recommendations (NVIDIA A100 or L20), and instructions for installing required third-party libraries and compiling \sys{} kernels.
    \item \textbf{Docker support} for reproducible deployment, including editable mount path and data path configurations, and build/run scripts.
    \item \textbf{Execution scripts} for reproducing our throughput and latency results using \texttt{demo.py}, which supports different model paths, speculative token lengths, and batch sizes.
    \item \textbf{Pretrained models} hosted on Huggingface for \sys{}'s Llama-3-8b-instruct model used in our vLLM experiments.
    \item \textbf{Notes and caveats} describing limitations of our current implementation (\textit{e.g.}, not optimized for all GPU types, partial vLLM integration, cold start auto-tuning delay).
    \item  \textbf{License and Intended Use:} We confirm that all third-party artifacts used in this work (\textit{e.g.}, vLLM, Huggingface-hosted models) were accessed and used in accordance with their licenses (Apache License 2.0) and intended research purposes. Our implementation of \sys{} is released under the Apache License 2.0 and is explicitly intended for academic and non-commercial use. Users are instructed to obtain such resources directly from their original providers and to comply with the corresponding terms of use.

\end{itemize}

All artifacts are accompanied by a \texttt{README.md} file that details the usage and experimental instructions. The code is released under an anonymous GitHub repository to ensure reproducibility.

\subsection{Understanding FP16 vs. W4A16 Performance in Main Results}\label{app:fp16}

\begin{figure}[h]
    \centering
    
    \includegraphics[width=0.48\textwidth]{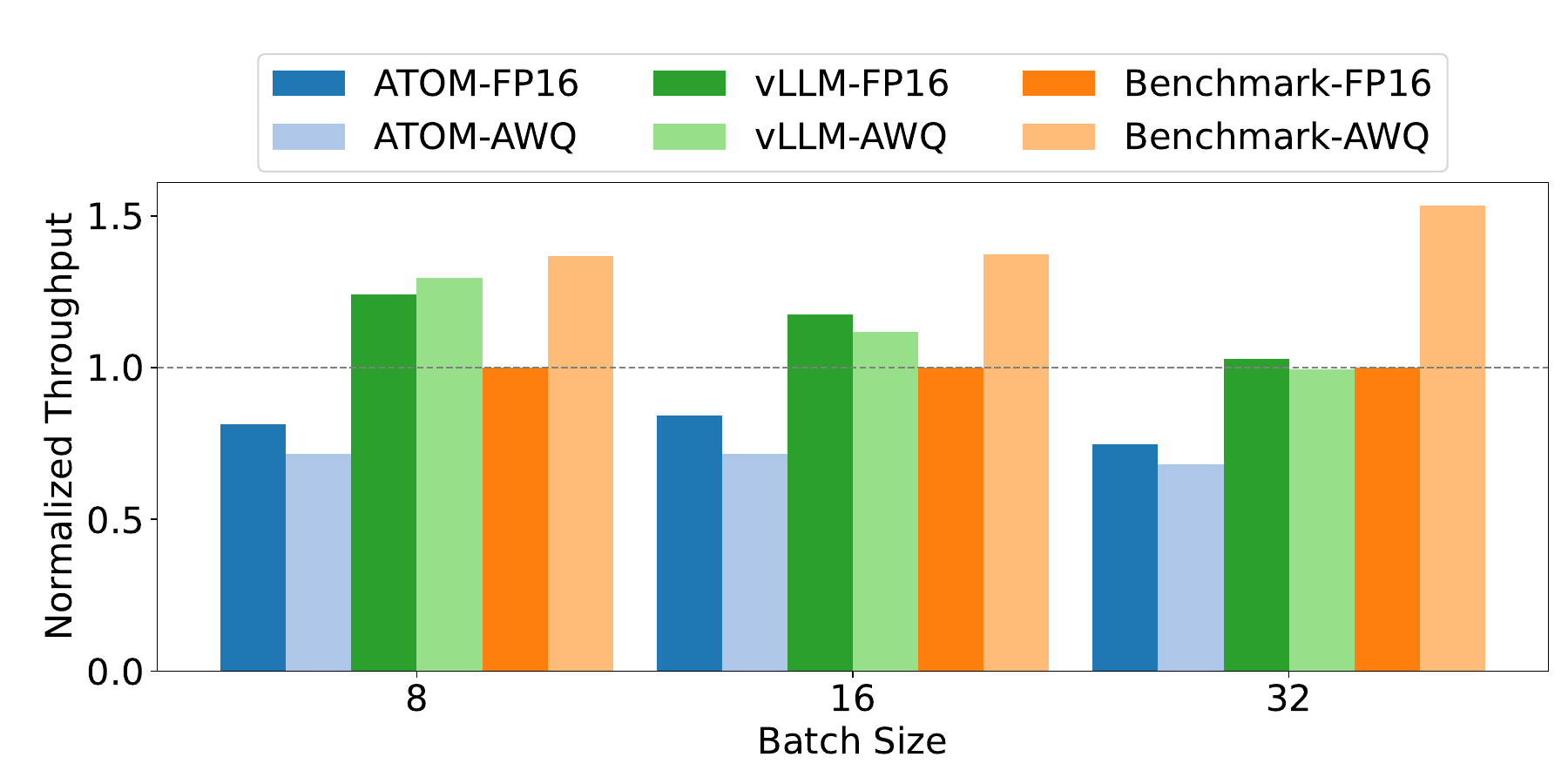}
    \caption{Normalized throughput of FP16 and AWQ implementations across batch sizes 8, 16, and 32. Atom-FP16 and Atom-AWQ use Atom’s system with FlashInfer and continuous batching. Benchmark-FP16 and Benchmark-AWQ are from AutoAWQ with optimized AWQ kernels and Flash-Attention, using a dummy benchmark method. vLLM-FP16 and vLLM-AWQ are implemented in vLLM. The generation length is set to 512.}
    \label{fig:fp16_vs_awq}
\end{figure}

While W4A16 quantization (\textit{e.g.}, AWQ) is often expected to outperform FP16 in small-to-medium batch sizes due to its design for improved efficiency in weight-only quantization~\cite{awq}, our main results consistently show FP16 surpassing W4A16 across various implementations, including vLLM, Atom’s system, and even Hugging Face’s official benchmarks\footnote{huggingface.co/docs/transformers/main/quantization/awq}. This discrepancy may lead some readers to question the relative performance, as implementation-specific factors such as device characteristics, kernel optimizations, and system engineering significantly influence outcomes. We conducted a complementary experiment to elucidate this phenomenon, illustrated in Figure~\ref{fig:fp16_vs_awq}, comparing FP16 and W4A16 under different implementation settings.

In this experiment, Atom-FP16 and Atom-AWQ are derived from Atom’s system implementation, following the end-to-end benchmark settings of our main experiments (Section 4), with FlashInfer integrated into a Punica-style serving system to support continuous batching~\cite{orca}. Conversely, Benchmark-FP16 and Benchmark-AWQ are sourced from the AutoAWQ repository, leveraging an optimized AWQ kernel and Flash-Attention, but employing a classic dummy benchmark method (directly invoking the model for the context length duration)\footnote{github.com/casper-hansen/AutoAWQ}. Additionally, we include vLLM-FP16 and vLLM-AWQ, implemented in the vLLM framework (Commit 9a7c3a0, Appendix A.4), to provide a broader perspective.

Figure~\ref{fig:fp16_vs_awq} reveals distinct performance trends across batch sizes of 8, 16, and 32. In Atom’s implementation, FP16 consistently outperforms AWQ across all batch sizes, aligning with our main results. However, in the AutoAWQ dummy benchmark, AWQ exhibits superior throughput, reversing the trend. In vLLM, AWQ slightly outperforms FP16 at a batch size of 8, but FP16 surpasses AWQ at batch sizes of 16 and 32. These variations underscore the impact of system implementation and kernel optimization on relative performance. Since our main experiments strictly adhere to Atom’s setup for fair and rigorous comparisons (Sec.~\ref{sec: exp}), the observed speed disparity between FP16 and W4A16 due to implementation differences does not undermine the validity of our claims regarding \sys{}’s performance.

\section{Evaluation Datasets}\label{app:datasets}

We comprehensively evaluate the performance of \sys{}, covering language modeling, commonsense reasoning, mathematical reasoning, code and chatbot.
Each dataset is briefly introduced below, along with the associated sampling strategy for enhanced efficiency.

\paragraph{Language Modeling and Commonsense Reasoning.}

\begin{itemize}
    \item \textbf{WikiText-2}~\citep{wikitext2}: A language modeling dataset comprising over 100 million tokens extracted from high-quality Wikipedia articles.
    \item \textbf{PIQA}~\citep{piqa}: A benchmark for physical commonsense reasoning, focusing on questions about everyday physical interactions.
    \item \textbf{Winogrande}~\citep{ai2:winogrande}: A dataset of 44,000 problems inspired by the Winograd Schema Challenge, designed to test commonsense reasoning with reduced linguistic biases.
\end{itemize}

\paragraph{Mathematical Reasoning}

\begin{itemize}
    \item \textbf{GSM8K}~\citep{gsm8k}: A collection of 8,500 linguistically diverse grade school math problems that require multi-step reasoning using basic arithmetic operations.
    \item \textbf{MATH}~\citep{MATH}: A dataset of 12,500 challenging competition-level math problems, each accompanied by detailed step-by-step solutions.
\end{itemize}

\paragraph{Code}

\begin{itemize}
    \item \textbf{MBPP}~\citep{MBPP}: A benchmark of approximately 1,000 beginner-level Python programming problems, each with a task description, solution code, and automated test cases.
    \item \textbf{HumanEval}~\citep{HUMAN_EVAL}: An evaluation set of 164 original programming problems used to assess functional correctness in code synthesis from docstrings.
\end{itemize}

\paragraph{Chatbot}

\begin{itemize}
    \item \textbf{ShareGPT52K}~\citep{sharegpt52k}: This dataset comprises approximately 52,000 conversations collected via the ShareGPT API before it was discontinued. The dataset captures both user prompts and the corresponding responses from OpenAI's ChatGPT, providing insights into human-AI dialogue dynamics.

    \item \textbf{LMsys-chat-1M-1K}~\cite{lmsys}: Gathering one million authentic conversations with 25 leading large language models (LLMs), this dataset was sourced from over 210,000 unique IP addresses interacting with the Vicuna demo and Chatbot Arena websites. 
\end{itemize}

\section{Datasets Sampling}

To construct our test sets, we randomly sampled from the original datasets using \textit{torch.sample} with a fixed seed of 42, and constructed them as prompt instructions following the Open-Instruct templates~\citep{wang2023far}. The sample sizes for each dataset are as follows:

\begin{itemize}
\item \textbf{Fidelity Evaluation:}
\begin{itemize}
\item \textbf{WikiText-2}: All samples
\item \textbf{PIQA}: 500 samples
\item \textbf{Winogrande}: 500 samples
\item \textbf{GSM8K}: 200 samples
\item \textbf{MATH}: 700 samples, balanced by selecting 100 questions from each of the seven distinct question types
\item \textbf{MBPP}: 200 samples
\item \textbf{HumanEval}: All samples
\end{itemize}

\item \textbf{Acceleration Evaluation:} 100 samples from each dataset, maximum output length set to 200 tokens.

\end{itemize}

\section{Related Work}\label{sec:appendix_related}

\textbf{Quantization} is a common technique for deploying LLMs on resource-limited scenarios. 
Broadly, recent quantization algorithms can be classified into two categories: weight-only W4A16 and weight-activation joint W4A4.
Notably, AWQ (W4A16)~\citep{awq} redistributes the quantization burden by scaling salient weight channels to protect them from degradation. In contrast, W4A4 aggressively quantizes activations to leverage low-precision hardware for improved speed at the cost of model quality degradation. To address this challenge, Atom~\citep{Atom} proposes reordering outlier channels in the activation through offline profiling. Similarly, QuaRot~\citep{quarot} employs Hadamard matrices to apply computational invariance on weights. Despite these advancements, our observations indicate that W4A4 methods still exhibit substantial degradation compared to weight-only quantization approaches across multi-step reasoning tasks. Works such as W4A8~\citep{qserve} and adaptive quantization~\citep{zhao2024llmpqservingllmheterogeneous,Dong2024QAQQA} seek to identify an optimal trade-off point. However, these methods struggle to fully preserve the generation quality associated with higher precision. 

\textbf{Speculative Decoding} leverages a draft model to generate candidate tokens, which are then validated by a target model \citep{leviathan2023fastinferencetransformersspeculative}. Recent research has primarily focused on improving the acceptance rate and generation speed of candidate tokens. SpecInfer \citep{spec_infer} introduces a boost-tuned small language model to generate candidate tokens in tree structures, enabling single-pass verification. In contrast, EAGLE \citep{eagle} adopts an aggressive pruning strategy for the draft model's architecture, allowing penultimate layer feature prediction with minimal computational overhead. Self-speculative decoding, a subset of this technique, employs a single model for both draft generation and verification. LayerSkip \citep{layerskip} introduces a training methodology for early exit with layer drop, subsequently verifying partially generated tokens through full model inference. Medusa \citep{Medusa} augments the original LLM with additional heads atop the final hidden state while relaxing the acceptance policy. However, these approaches inevitably require retraining of the original model, which can be computationally expensive and time-consuming. We further demonstrate their deficiency in batched serving under quantization scenario. 

\textbf{Tree-Structured Drafting.}
Tree-structured drafting~\citep{sequoia, spec_infer, eagle, eagle2} is a widely adopted technique to improve acceptance rates in speculative decoding. Instead of generating a single draft token at a time using the draft model $\mathcal{M}_d$, $\mathcal{M}_d$ select $k$ (\textit{i.e.}, top-k) tokens and infers $\gamma$ times to form a draft tree with depth $\gamma$. The target model $\mathcal{M}$ then verifies the tokens using masked tree attention, with the exact verification strategy depending on the sampling method~\citep{Medusa, spec_infer}. Under greedy sampling, the highest-probability token at each position is selected, forming the longest branch with the common prefix as the accepted sequence. In this tree, the root node presents the past sequence, any token $t_i$ has a path from the root node $r$ to $t_i$, denoted by $Path(r, t_i)$, consisting of ancestors $a_1, \ldots, a_j$.

\section{Supplementary Figures and Tables}
\label{sec:supp_figures}

To improve the clarity of the main text and streamline presentation, we provide additional visualizations and ablation results related to \sys{} in this appendix. 
These include supporting data referenced in the main body and additional experiments.

\begin{table}[H]
\centering
\caption{Ablation study comparing acceptance rates (\%) across base quantization methods using \sys{}.}
\label{tab:acceptance_rate_for_different_quant_methods}
\resizebox{\linewidth}{!}{%
\begin{tabular}{lccc}
\toprule
\makecell{\textbf{Quantization Method}}
& \textbf{ShareGPT}
& \textbf{MATH (4-shot)}
& \textbf{MBPP (0-shot)} \\
\midrule

\multirow{1}{*}{Atom}
& 83.8 & 89.4 & 88.6 \\

\multirow{1}{*}{QuaRot}
& 81.6 & 88.9 & 85.4 \\

\bottomrule
\end{tabular}%
}
\end{table}

\begin{table}[h!]
\centering
\caption{Ablation study comparing acceptance rates (\%) across large reasoning model and difficult reasoning tasks using \sys{}.}
\label{tab:model_acceptance_rates_ablation}
\resizebox{\linewidth}{!}{%
\begin{tabular}{@{}lccccc@{}}
\toprule
\textbf{Model} & \textbf{GPQA-Diamond} & \textbf{Super-GPQA} & \textbf{AIME} & \textbf{ARC} & \textbf{MMLU} \\ \midrule
\multirow{3}{*}{\makecell[l]{Llama3-8b\\-Instruct}}
& 94.1 & 96.5 & 96.1 & 92.6 & 92.4 \\
\cmidrule(l){2-6}
& \textbf{OpenBookQA} & \textbf{RACE} & \textbf{SQuAD v2} & \textbf{TruthfulQA} & \textbf{HellaSwag} \\
& 92.6 & 94.2 & 95.0 & 92.0 & 91.7 \\
\cmidrule(l){2-6}
& \textbf{HumanEval} & \textbf{LAMBADA} & \textbf{Social IQa} & \textbf{Avg.} & \\
& 87.5 & 89.6 & 93.8 & \textbf{92.93} & \\ \midrule
\multirow{3}{*}{\makecell[l]{DeepSeek-R1-Distilled\\-QWen14B}}
& 96.2 & 96.0 & 97.9 & 90.4 & 90.8 \\
\cmidrule(l){2-6}
& \textbf{OpenBookQA} & \textbf{RACE} & \textbf{SQuAD v2} & \textbf{TruthfulQA} & \textbf{HellaSwag} \\
& 90.5 & 92.8 & 92.8 & 88.6 & 96.7 \\
\cmidrule(l){2-6}
& \textbf{HumanEval} & \textbf{LAMBADA} & \textbf{Social IQa} & \textbf{Avg.} & \\
& 96.7 & 94.3 & 91.7 & \textbf{93.49} & \\ \bottomrule
\end{tabular}%
}
\end{table}

\begin{table}[H]
\centering
\caption{Performance results for the reasoning model DeepSeek-R1-Distilled-QWen14B. We follow the settings of our main experiments and set the batch size (bs) to 16.}
\label{tab:deepseek_performance}
\resizebox{\linewidth}{!}{%
\begin{tabular}{l|ccccccc}
\hline
\textbf{Dataset} & \textbf{GSM8K} & \textbf{MATH} & \textbf{MBPP} & \textbf{HUMANEVAL} & \textbf{SHAREGPT} & \textbf{LMSYS} & \textbf{AVG.} \\ \hline
W4A16 (tokens/s) & 139.38 & 203.52 & 200.84 & 216.54 & 194.31 & 194.22 & 191.47 \\
QSpec (tokens/s) & 171.56 & 282.17 & 278.74 & 292.75 & 259.48 & 246.07 & 255.13 \\
Speedup & 1.23 & 1.39 & 1.39 & 1.35 & 1.34 & 1.27 & 1.33 \\ \hline
\end{tabular}%
}
\end{table}

\section{AI Assistance Statement}

We used AI tools (\textit{e.g.}, ChatGPT) exclusively for language polishing and LaTeX formatting. No part of the core research, including ideation, experimental design, or analysis, was generated by AI tools. All scientific contributions are the sole work of the authors.

\end{document}